\documentclass[final,3p,times]{elsarticle}  

\usepackage{amssymb}

\usepackage[utf8]{inputenc} 
\usepackage[T1]{fontenc}    
\usepackage{hyperref}       
\usepackage{url}            
\usepackage{booktabs}       
\usepackage{amsfonts}       
\usepackage{nicefrac}       
\usepackage{microtype}      
\usepackage{graphicx}       
\usepackage{amsmath}        
\usepackage{commath}        
\usepackage{floatrow}       
\usepackage{subfig}         
\usepackage{svg}            
\usepackage[space]{cite}    
\usepackage{soul}           
\usepackage[ruled]{algorithm2e}  
\usepackage{setspace}       

\journal{Journal of Biomedical Informatics}

\begin{document}
\onehalfspacing  

\begin{frontmatter}



\title{Interpreting a Recurrent Neural Network's Predictions of ICU Mortality Risk}


\author[1]{Long V. Ho}

\ead{loho@chla.usc.edu}
 
\author[1]{Melissa Aczon}
\ead{maczon@chla.usc.edu}

\author[1]{David Ledbetter}
\ead{dledbetter@chla.usc.edu}

\author[1]{Randall Wetzel}
\ead{rwetzel.chla.usc.edu}

\address[1]{The Laura P. and Leland K. Whittier Virtual Pediatric Intensive Care Unit\\
    Children's Hospital Los Angeles, 4650 Sunset Blvd, Los Angeles, CA 90027, United States\\
}

\begin{abstract}
Deep learning has demonstrated success in many applications; however, their use in healthcare has been limited due to the lack of transparency into how they generate predictions. Algorithms such as Recurrent Neural Networks (RNNs) when applied to Electronic Medical Records (EMR) introduce additional barriers to transparency because of the sequential processing of the RNN and the multi-modal nature of EMR data. This work seeks to improve transparency by: 1) introducing Learned Binary Masks (LBM) as a method for identifying which EMR variables contributed to an RNN model's risk of mortality (ROM) predictions for critically ill children; and 2) applying KernelSHAP for the same purpose. Given an individual patient, LBM and KernelSHAP both generate an attribution matrix that shows the contribution of each input feature to the RNN's sequence of predictions for that patient. Attribution matrices can be aggregated in many ways to facilitate different levels of analysis of the RNN model and its predictions. Presented are three methods of aggregations and analyses: 1) over volatile time periods within individual patient predictions, 2) over populations of ICU patients sharing specific diagnoses, and 3) across the general population of critically ill children.
\end{abstract}


\end{frontmatter}


\section{Introduction}
\label{section:intro}
Deep learning has demonstrated promising results in a wide variety of healthcare domains including radiology \citep{cheng2016computer, cicero2017training, gulshan2016development}, oncology \citep{kooi2017large, esteva2017dermatologist, liu2017detecting}, and intensive care \citep{choi2016doctor, suresh2017clinical, lipton2015learning, aczon2017dynamic, winterwlst}. This is due to the increasing availability of large clinical datasets such as the Electronic Medical Records (EMR) \citep{henry2016adoption} and advances in computing technology that enable practical training of deep learning models \citep{esteva2019guide}. The promise of deep learning is its ability to learn complex interactions directly from high-volume, high-dimensional, and multi-modal data without the need for hand selecting and engineering features specific to a modeling technique or problem \citep{lecun2015deep}. Unfortunately, this flexibility comes at a price: a model with millions of parameters and hundreds of operations, opaquely optimized from large complex datasets. As a result, how a particular input feature contributes to or affects a prediction is not immediately obvious.

This lack of transparency, especially in clinical settings where decisions may be lifesaving, has inspired research efforts to interpret these highly accurate and complex models \citep{samek2017explainable}. Despite this growing interest, \textit{interpreting} a model remains a nebulous concept and is usually defined specifically for the problem and application of the model \citep{doshi2017towards}. Consequently, methods for interpreting deep learning models are very diverse in method and purpose; for example, aggregating and visualizing the model's neuron activations to extract concepts learned \citep{olah2017feature} or using a simpler model such as a decision tree to approximate the predictions of the original model and interpreting the simpler model as a proxy \citep{che2016interpretable}. The general goal of methods for interpreting is to understand the model's \textit{decision making process}. In this work, we use a simplified definition for \textit{interpreting} a model -- that is understanding which inputs contributed to the model's predictions.

In particular, we are interested in determining which input features contributed to the predictions of a previously well described recurrent neural network (RNN) model that uses electronic medical data to continuously assess the status of a critically ill child based on their risk of mortality (ROM) in a pediatric intensive care unit {\citep{aczon2017dynamic}}. The ability to determine how input features impact these predictions is important for several reasons.

First, it may provide useful information for clinical intervention. ICU mortality predictions for an individual patient serve as a proxy for a \mbox{child’s} severity of illness (SOI) {\mbox{\citep{pollack2016pediatric,tasker2016severity,pollack2016severity,leteurtre2015daily}}}. They have intrinsic value, but understanding which of the child's features underlie the acute changes of their state, as indicated by a change in the predicted SOI, would add further value. If the clinician already knows this information, then it provides corroboration and trust, and the clinician would know what to do with it (e.g. blood pressure is important, therefore administer a therapy to optimize blood pressure). If not, then it may propel further investigations that otherwise may have been overlooked without the model's prediction and interpretation. 

Second, it facilitates an environment in which users can interact with the model and learn its strengths and weaknesses. Users can then compare the extracted input contributions with their own clinical experience {\mbox{\citep{molnar2019,doshi2017towards}}}. 

Third, understanding which or how inputs contributed to predictions can be used to improve the model. Combining this understanding with clinical knowledge can identify when the model improperly leverages information. Determining and correcting such characteristics are especially important in deep learning models where parameters are optimized to large biased datasets commonly found in healthcare. The bias comes from the observational nature of healthcare data, where counterfactual events do not occur. For example, if a drug is given only to terminally ill patients during end-of-life withdrawal of support, a model may inadvertently leverage the use of the drug as an indicator of mortality, which would contradict the intended purpose of the model, e.g. to find meaningful features that can be changed to improve survival probability.

The above reasons motivated the following primary goals of this work, which also describe this paper's contributions to the still limited but growing body of literature on the methods of interpreting deep learning models applied to EMR:

\begin{itemize}
    \item Introduce the Learned Binary Mask (LBM), a new occlusion-based method, to identify which inputs contributed to the predictions of a many-to-many RNN model that continuously generates ROM scores for an individual child from multi-modal time series EMR data. The LBM is able to manage the mixed data modalities in Electronic Medical Records.
    \item Modify KernelSHAP to make it compatible with a many-to-many RNN model for risk of ICU mortality whose inputs are multi-modal EMR.
    \item Use both the LBM and KernelSHAP to compute attribution matrices across all individual patients. Aggregate the attribution matrices in various ways for different level of analysis of the RNN model and its predictions: 1) within volatile time periods in individual patient predictions, 2) across cohorts of children diagnosed with specific diseases (sepsis and brain neoplasm), and 3) across the general population of critically ill children. These use cases of interpreting the {\mbox{many-to-many}} RNN {\mbox{model’s}} predictions with the LBM and Kernel SHAP emphasize the importance of understanding the mathematics and assumptions of each method when applied to real data to understand the analysis of the results.
    \item Introduce a novel matrix representation which reflects hour-to-hour \mbox{\emph{state changes}} that a patient undergoes during their stay in the ICU. This matrix is the input to the RNN model that generates dynamically evolving predictions of the patient's ICU mortality risk (severity of illness). The state change representation enables use of the LBM and KernelSHAP.
\end{itemize}

We emphasize that this work presents the LBM and KernelSHAP as complementary, not competing, methods. Their formulations and purposes are different; therefore, one method's results should not be regarded as better or more accurate than the other. Importantly, evaluating methods for model interpretation is inherently qualitative because there are no \emph{hard truths} against which to compare the outputs of such methods. The evaluations rely on clinical insights and experience, which are not necessarily quantifiable. Nevertheless, these qualitative analyses and evaluations are important for the reasons stated earlier. Finally, note that this study is not about feature selection for model development.

\section{Related Works}
\label{section:related_works}
Using both RNNs \textit{and} multi-modal EMR data poses challenges to current interpretation methods: RNNs introduce complexities associated with the time dimension while EMR data complicate comparisons among features that have different distributions and clinical meaning. Many methods for interpreting deep learning models rely on sensitivity analysis which measures feature attributions by analyzing how the prediction changes when inputs are perturbed \citep{Guidotti:2018:SME:3271482.3236009}. These methods are limited to single-modal inputs such as images or text in which changes in inputs and outputs can be compared readily among features. In contrast, data in EMRs contain an eclectic collection of data modalities which cannot be trivially compared, including continuous physiology (heart rate), categoricals (Glasgow Coma Score), binary (cultures), and unstructured texts (clinical notes) \citep{goldstein2017opportunities}. In addition, many methods rely on special visualizations for presenting the extracted information, e.g. heatmaps highlighting important pixels in an image. Such visualizations for EMR data and RNNs can be intractable because even a single patient's data can contain hundreds of different measurements and thousands of time steps. 

One approach to address these issues has been to adapt the problem to current methods of interpretability. For example, Rajkomar et al. \citep{rajkomar2018scalable} converted the problem into the familiar problem of text processing by tokenizing the EMR data as single-sensor text sequences and interpreting the RNN model with methods developed for natural language processing. Another way to leverage existing interpretability techniques is to use mimic learning wherein a simpler model approximates the complex model. This approach was taken by Che et al. \citep{che2016interpretable} who approximated their RNN-EMR model with a gradient boosted trees model (GBM) trained to predict the RNN's predictions; interpreting the GBM was a proxy for interpreting the original RNN-EMR model. Such methods often require non-trivial manipulation of the data or model, and this process introduces additional layers of complexity that further muddles interpretation.

Other algorithms that use RNNs and EMR data are interpreted by embedding certain components of the algorithm with interpretable constituents. Cerna et al. \citep{cerna2019interpretable} aggregated the response from models trained specifically on individual modalities of the EMR and interpreted the weight of a final linear layer as contributions from each of the mixed modalities. Choi et al. \citep{choi2016retain} and Zhang et al. \citep{zhang2018patient2vec} used attention RNNs to interpret their models, modifying RNNs with an attention component that imposes an explicit attribution of the inputs to the outputs via weights. Such methods only interpret \textit{parts} of the deep network and require complex visualization techniques to distill the information. Furthermore, Poursabzi-Sangdeh et al. \citep{poursabzi2018manipulating} found that more transparent models, when compared to the same models that were presented as black-boxes, had no benefits in application and actually had detrimental effects due to information overload.

Another approach has been to use \textit{explanation models}, which are interpretable meta-models trained in addition to the original model. Compared to other interpretability methods, explanation models are constructed to investigate specific properties of the original model (e.g. rotational invariance of the model). Both KernelSHAP and LBM fall into this category. Explanation models are specific to a particular problem. To the authors' knowledge, the only publication that uses explanation models on RNNs and EMR data is in Suresh et. al \citep{suresh2017clinical}, which examined the impact of features by comparing differences in predictions when individual features were included or excluded. 
Further, the authors are not aware of KernelSHAP applications to RNN models using EMR data. This is likely due to the limitations of the current KernelSHAP implementation {\citep{kshapgithub}}. To facilitate discussion, the formulation of KernelSHAP and some implementation choices made for this study are described in Section \ref{section:methods_kshap}.

The LBM extends the methodology of Fong \& Vedaldi \citep{fong2017interpretable} for interpreting CNNs and images to RNNs and EMRs. In Fong \& Vedaldi, the
fundamental concept is to find a mask that identifies which set of pixels of an image removes evidence for being in the class of interest. Similarly, the LBM method finds a \textit{binary} mask, instead of a real-valued mask, that identifies which set of input features when set to zero removes evidence for mortality. The LBM's significant departure from established occlusion-based methods is its \textit{binary} constraint on the mask, which is nontrivial to obtain in practice but essential for comparing the contributions of multi-modal features in the EMR data.

The aforementioned explanation methods generate what are known as local explanations: for each individual prediction, they compute the contribution of each input feature. These local explanations can be aggregated over different predictions to provide global insights. For example, Lundberg et al.{\citep{lundberg2019explainable}} aggregated local explanations across entire datasets to compute traditional model feature importance, revealing the average contribution of features and avoiding problems associated with traditional global explanations. Similarly, we extended this process by aggregating and normalizing across sub-populations. This was used to identify which features had relatively high contributions to mortality risk predictions in different diagnosis groups and the general ICU population.

\section{Methods}
\label{section:methods}
\subsection{Data}
This study used de-identified EMR data collected in the Pediatric Intensive Care Unit (PICU) from Children's Hospital Los Angeles (CHLA). The CHLA Institutional Review Board (IRB) reviewed the study protocol and waived the need for IRB approval. The data consisted of 9855 PICU encounters (7358 patients) from 2009 to 2017 ($4\%$ mortality rate), where an encounter is defined to be a patient's contiguous stay in the ICU. A patient can have multiple ICU encounters. Data for each patient encounter included irregularly sampled physiologic observations, laboratory results, drugs, and interventions (e.g. intubation parameters). Also collected were the patient's demographics, diagnoses, and outcomes (e.g. ICU mortality). The encounters were partitioned into training ($60\%$), validation ($20\%$), and test ($20\%$) sets, where the partitioning was done such that all encounters from a single patient belonged to only one of the sets. Statistics of the datasets are included in \ref{section:appendix:a}.\\

\begin{figure*}[htbp!]
\centering
\includegraphics[width=1.0\textwidth]{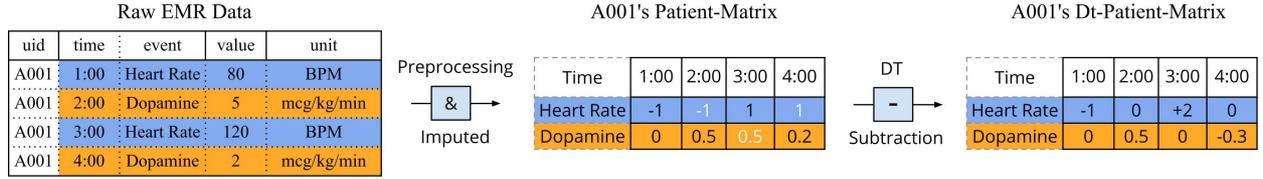}
\caption{Pre-processing from raw EMR data to the  \textit{dt-patient-matrix}. Raw EMR Data is converted to a \textit{patient-matrix} by pivoting the data to wide-format, imputing missing data with forward fill, converting physiologic variables to z-scores using means and standard deviations (Heart Rate with $(\mu, \sigma)=(100, 20)$), and normalizing exogenous variables such as drugs to [0, 1] using minimum and maximum values (Dopamine with $(\mbox{min}, \mbox{max})=(0, 10)$) computed from the training set. Note the normalization parameters used here were chosen for illustrative purposes. Finally, data from the \textit{patient-matrix} is converted to the \textit{dt-patient-matrix} by subtracting values from subsequent time-steps. Text colored in white (in A001's Patient-Matrix) indicates imputed data.}
\label{fig:data_preproc}  
\end{figure*}

\subsubsection{Patient Representation: The ``patient-matrix''}
Pre-processing steps converted the EMR data to matrices that facilitated machine learning while maintaining an interpretable patient representation. In collaboration with physicians, lower and upper limits for all variables were defined, and the entries were curated to remove observations not compatible with life. Values from different methods of measurements of the same variable were combined into a single variable when medically appropriate (e.g. invasive and non-invasive blood pressure). Physiologic variables such as labs and vitals were converted to z-scores using means and standard deviations computed from the \textit{training set}. In the z-score space, the features associated with physiologic variables represent how \textit{far} the patient's physiologic measurements are from the ICU \textit{averages}, and this distance is measured in terms of standard deviations. Exogenous variables such as interventions and drugs were normalized to values from 0 to 1 using the lower and upper limits of values/dosages computed from the \textit{training set}. Consequently, features associated with exogenous variables in the transformed space represent a percentage of therapies administered in the ICU, with 0 indicating no therapy and 1 denoting maximum therapy possible within the dataset. 

Each patient's data were forward-filled and re-sampled to hourly observations, then pivoted to form a sparse $N \times T$ \textit{patient-matrix}, where $T$ is the number of distinct timestamps (varied across encounters), and $N$ is the total number of input features (see Appendix {\ref{section:appendix:a}} for a list of the $N=398$ features). The training set mean was used to impute a physiologic or laboratory variable at all times prior to its first recorded value or at all times of the episode if it had no recorded value for the entire episode. Any missing measurement of a treatment indicates actual absence of that treatment; therefore it was set to zero. Note that patient diagnoses were not included as input features. A single column of this matrix contains measurements from all $N$ features at a single time point, while a single row contains measurements of a single feature from all $T$ time points. A similar matrix representation of patient encounter data, but without the hourly re-sampling, was used in previous work \citep{aczon2017dynamic, ho2017dependence, laksana2020impact,winterwlst}.\\

\subsubsection{Patient Representation: The ``dt-patient-matrix''}
\label{section:data:dt}
To facilitate the application of both LBM and KernelSHAP, we further transformed the \mbox{\textit{patient-matrix}} to represent the changes between each time step: the \mbox{\textit{dt-patient-matrix}}. The \mbox{\textit{patient-matrix}} was transformed to \textit{dt-patient-matrix} with these steps: the first column of the \textit{dt-patient-matrix} is exactly the same as the first column of the original patient matrix, which represents the patient's state relative to the ICU population encapsulated in the training set. Values in subsequent columns of the \textit{dt-patient-matrix}, each indexed by $\tau$, were obtained by subtracting the patient state at time $\tau-1$ from the state at time $\tau$. Figure \ref{fig:data_preproc} illustrates the pre-processing steps from raw EMR data to the \textit{dt-patient-matrix}. The \textit{dt-patient-matrix} captures the hour-to-hour changes -- physiologic and therapeutic -- that a patient undergoes during an ICU encounter.

Because sparsely measured features such as laboratory tests and rare treatments were included, pivoting the long-format EMR data to the original \textit{patient-matrix} representation introduced a sparse matrix with over 94\% of elements set to zero (measured across the $398$ selected observations). In the \emph{dt-patient-matrix} representation, the forward-filled values became zeros, indicating either no change from the population average (if the element is in the first column) or no change from the individual's previous state (subsequent columns). This representation is consistent with the collection of data in clinical practice: new observations are typically recorded during state changes and are assumed to be the previous value until a new recording or entry is made \citep{donabedian1966evaluating, donabedian1988quality}. 

Importantly, this state change representation provides significant and distinctive advantages when using the LBM or KernelSHAP. The \textit{dt-patient-matrix} removes ambiguities when occlusion-based methods of interpretation are used. To ``delete'' (i.e. set to zero) an element precisely means to have no change in that variable. This representation also facilitates practical use of KernelSHAP:  the zero-valued elements of the dt-patient-matrix define a set of ``missing'' features, and having this set bypasses expensive computational steps otherwise required. These points are further discussed in Sections {\ref{section:methods_lbm}} and {\ref{section:methods_kshap}}.

\subsection{RNN Model for ICU Mortality}\label{section:methods:rnn}
\begin{figure*}[htbp!]
\centering
\includegraphics[width=0.9\textwidth, height=0.3\textheight]{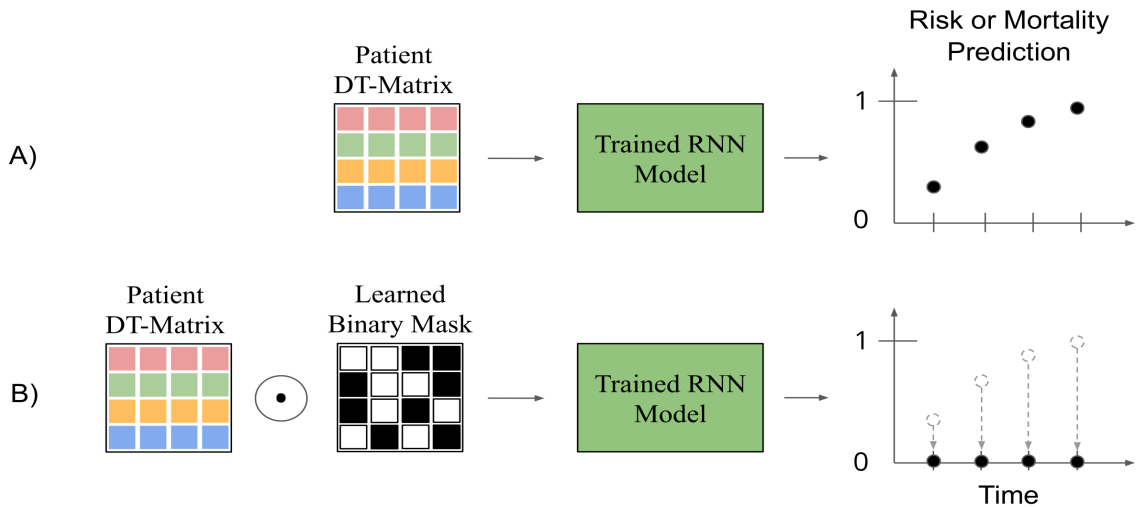}
\caption{A) Every hour after ICU admission, a many-to-many RNN model acquires new patient data (single column of the patient dt-matrix) and generates a mortality prediction (Section {\ref{section:data:dt}}). The Patient DT-matrix is color coded to denote the multi-modal nature of the input data: vitals in blue, labs in green, interventions in orange, and drugs in red.  B) The Learned Binary Masks algorithm, detailed in Section {\ref{section:methods_lbm}} and {\ref{section:appendix:b}}, computes a matrix of binary-valued weights such that when these weights multiply the input data, the RNN's risk of mortality predictions go to zero at all time steps. The zeros in the binary matrix identify which of the input Patient DT-matrix elements made the ROM predictions go to zero. This is analogous to occlusion-based methods in images that find the set of pixels such that when they are `deleted' or masked, the class probability goes from non-zero to zero {\citep{fong2017interpretable}}.}
\label{fig:lbm_conceptual}  
\end{figure*}

An RNN model using the dt-patient-matrix as input was trained to predict ICU mortality. Given a patient encounter, the model generates a risk of mortality (ROM) each time it receives new data (i.e. a single column from the \emph{dt-patient-matrix}); see Figure {\ref{fig:lbm_conceptual}A}. The model is composed of three stacked Long Short-Term Memory (LSTM) \citep{hochreiter1997long} layers with hidden units 128, 256, and 128 respectively and a final logistic regression layer for classification. Each layer's weights were initialized using Glorot uniform \citep{glorot2010understanding} and optimized using RMSprop \citep{tieleman2012lecture}. Using an initial learning rate of $1e^{-4}$ and mini-batch size of 128, the model was trained to minimize the binary cross-entropy between the model's predicted ICU-mortality risk and the patient's true mortality response, repeated for every time-step. Performance on the validation set was evaluated after every epoch (a full cycle through the training set), and the best performing weights were saved. If the validation set's binary cross-entropy did not decrease after 15 epochs, learning rate was decreased by a factor of 5 and terminated after 2 reductions. Model regularization included a 20\% dropout of the output of each layer and an L2 penalty of $1e^{-5}$ against each layer's weights. The Python package Keras \citep{chollet2015keras} and the TensorFlow \citep{tensorflow2015-whitepaper} backend were used to construct and train the model. Training the full model using a Titan V GPU took approximately 6 hours to complete. 

The RNN model's performance was evaluated by computing the area under the ROC curve (AUC) across the hold-out test set at various times: 1, 3, 6, 9, 12, -12, -9, -6, -3, -1, where a positive number indicates hours since ICU admission and negative indicates hours until ICU discharge (or death). In this test formulation, data up until $t-1$ is given as input to generate a ROM prediction which is compared with the outcome at $t$. For example, data from $t=0$ to $t=11$ is used to evaluate the AUC at $t=12$. This is presented in Section \ref{section:results:rnn_perf}. Clinically used risk of mortality models, PIM2 \citep{slater2003pim2} and PRISM3-12 \citep{pollack1996prism}, were also evaluated on the same test set to provide comparators. Note that PIM2 and PRISM-3 are static models (using logistic regressions) and use data within 12 hours of ICU admission to generate a single prediction per patient encounter. Consequently, their performances can only be fairly compared with the RNN's $12^{th}$ hour predictions. To ensure proper comparison of AUCs of the RNN predictions at different hours, patients with less than 24 hours of ICU data were excluded from the AUC computations. The AUC was chosen as the metric for performance evaluation because it is not sensitive to class imbalance.  Preserving the class imbalance (96\% vs 4\%) during both model training and assessment is important because this imbalance would be present during actual deployment and would inform a full analysis of deployment benefits and costs.

\subsection{Learned Binary Masks}
\label{section:methods_lbm}
The goal is to find a binary mask for each input element in the dt-patient matrix such that when applied, the RNN model predictions go to zero. See Figure 2B for a conceptual illustration and \ref{section:appendix:b} for an algorithmic description. The reasoning for this formulation is analogous to the premise behind occlusion-based methods for interpreting convolutional neural networks that classify images: find the set of pixels such that when they are `deleted' or masked, the class probability goes from non-zero to zero {\citep{fong2017interpretable}}. 
The binary requirement stems from the multi-modal nature of EMR data which include real-valued, integer-valued, categorical, and binary features. Multiplying a binary or integer-valued feature by a fractional mask value can lead to a clinically unrealistic transformation. For example, multiplying the variable indicating whether or not a chest X-Ray was taken by a 0.5 mask has no meaning. Similarly, a fractional multiplier for Glasgow Coma Score, which can only take on integer values from 3 to 15, is not realistic. For generalizability (ie. to accommodate models that use potentially different sets of EMR variables), the method that generates the weights needs to be independent of the exact structure of the inputs. This dictates the binary nature of the resulting weights.

We focus on the RNN ICU mortality model described in Section {\ref{section:methods:rnn}}. Each patient encounter results in a matrix of binary masks indicating whether elements in the encounter's dt-patient-matrix elements contributed to the trajectory of ROM predictions for that encounter.  The formulation of the input dt-patient-matrix means that assigning a zero weight for a particular feature at a specific time step is equivalent to requiring no recorded change in that feature from the previous to the specified time step.

\subsubsection{Mathematical Formulation}
The RNN ICU mortality model and derivation of LBM are mathematically 
formalized to facilitate understanding of how the LBM determines feature attributions. The mortality model is a many-to-many recurrent neural network $f$ with trained parameters $\Theta$ that maps $\boldsymbol{x}_{1:T} \in \mathbb{R}^{N \times T}$ (\textit{dt-patient-matrix} with $T$ time points and $N=398$ features) to a $T$-length sequence of probabilities of mortality:
\begin{equation}
\label{equation:rnn_basic}
f(\Theta; {x}_{1:T}) = [y_1, y_2, \ldots, y_T] \equiv \boldsymbol{y}_{1:T} .
\end{equation}
At any single time, $\tau$, of an individual patient encounter, the notation $y_\tau=0$ corresponds to zero risk of mortality, while $y_\tau=1$ corresponds to 100\% risk of mortality. 

Let $\boldsymbol{0}_{1:T}$ denote a $T$-length sequence of zeros. If $\boldsymbol{y}_{1:T} \neq \boldsymbol{0}_{1:T}$, i.e. the prediction is not zero at every time step, then LBM will find a sparse, binary mask $\boldsymbol{M}_{1:T} \in \{0,1\}^{N \times T}$ that satisfies
\begin{equation}
\label{equation:lbm_basic}
 f(\Theta; \boldsymbol{M}_{1:T} \odot x_{1:T}) = \boldsymbol{0}_{1:T}, 
\end{equation}
where $\odot$ denotes the Hadamard product (element-wise matrix multiplication). This is illustrated in Figure {\ref{fig:lbm_conceptual}B}. If $\boldsymbol{M}$ solves Equation {\ref{equation:lbm_basic}}, then the locations of $\boldsymbol{M}$'s zeros correspond to the elements of the input dt-patient-matrix that must be zeroed for the predictions to go to zero. If the inputs identified by $\boldsymbol{M}$ are not zeroed (i.e. input $x^j_{\tau}$ retains its original non-zero value), then the ROM prediction at $\tau$ remains non-zero. Therefore, a solution with $M^j_{\tau}=0$ indicates that the recorded change in feature $j$ from time $\tau-1$ to time $\tau$ contributed to the non-zero ROM prediction at time $\tau$, i.e the recorded change provided evidence for mortality. 

The mask $\boldsymbol{M}^{*}_{1:T}$ that satisfies Equation \ref{equation:lbm_basic} can be found by leveraging the same fundamental mechanics used to train the neural network -- minimization of a regularized loss function through backpropagation. Mathematically, this is expressed as
\begin{equation}
\label{equation:m_loss_opt_1}
    \boldsymbol{M}^*_{1:T} = \underset{\boldsymbol{M}_{1:T}}{\textrm{argmin}} f\left(\Theta; \boldsymbol{x}_{1:T}\odot \boldsymbol{M}_{1:T} \right) + R(\boldsymbol{M}_{1:T}),
\end{equation}
where the first term, $f\left(\Theta; \boldsymbol{x}_{1:T} \odot \boldsymbol{M}_{1:T} \right)$, ensures that the mask causes the predictions to go to 0, while the second term, $R(\boldsymbol{M}_{1:T})$, imposes sparsity on the mask via $L_1$ regularization: $R(\boldsymbol{M}_{1:T}) = \lambda_1 \norm{1 - \boldsymbol{M}_{1:T}}_1$. This functional form for $R$ pushes many of the mask entries to unity, which is equivalent to minimizing the number of zero entries in the mask. This means that only those features that provided evidence for non-zero ROM, i.e. mortality, are selected. 

The non-differentiable nature of the binary constraint on $\boldsymbol{M}^{*}_{1:T}$ poses a challenge to optimization methods that leverage gradients such as backpropagation. A two-step process addresses this issue: (1) use backpropagation to solve Equation \ref{equation:m_loss_opt_1} for an intermediate non-binary mask, $\boldsymbol{m}_{1:T}\in [0, 1]^{N \times T}$; and (2) find another non-binary mask, $\boldsymbol{\eta}_{1:T}\in [0, 1]^{N \times T}$ to binarize $\boldsymbol{m}_{1:T}$. The final binary mask is given by $\boldsymbol{M}_{1:T} \equiv (\boldsymbol{m}_{1:T} > \boldsymbol{\eta}_{1:T})$.

The first mask $\boldsymbol{m}_{1:T}$ is found by applying backpropagation to a modified version of Equation \ref{equation:m_loss_opt_1}:
\begin{equation}
\label{equation:m_float_loss_opt_2}
    \boldsymbol{m}^*_{1:T} = \underset{\boldsymbol{m}_{1:T}\in [0, 1]^{N \times T}}{\textrm{argmin}} f\left(\Theta; \boldsymbol{x}_{1:T}\odot \boldsymbol{m}_{1:T} \right) + \lambda_1^1 \norm{1 - \boldsymbol{m}_{1:T}}_1 + \lambda_2^1 H \left( \boldsymbol{m}_{1:T} > 0.5, \boldsymbol{m}_{1:T} \right),
\end{equation}
where $H$ is the binary cross-entropy function, and $\lambda_1^1$ and $\lambda_2^1$ are regularization constants. The first two terms parallel those in Equation \ref{equation:lbm_basic}, while the third term, $\lambda_2^1 H \left( \boldsymbol{m}_{1:T} > 0.5, \boldsymbol{m}_{1:T} \right)$, encourages the values of $\boldsymbol{m^*}_{1:T}$ to be closer to 0 or 1, i.e. pushes $\boldsymbol{m^*}_{1:T}$ to be \textit{near-binary}. 
Next, let $\boldsymbol{m}_{1:T} \equiv \sigma (A \times \boldsymbol{z}_{1:T})$, where $A$ is a constant, $\sigma$ is the sigmoid function $\frac{1}{1 + e^{-x}}$, and $\boldsymbol{z}_{1:T} \in \mathbb{R}^{N \times T}$; and similarly define $\boldsymbol{m}^*_{1:T}$ from $\boldsymbol{z}^*_{1:T}$. Optimizing for $\boldsymbol{z}^*_{1:T}$ instead of $\boldsymbol{m}^*_{1:T}$ removes the $[0,1]$ range constraint during the backpropagation process.

Next, the threshold mask $\boldsymbol{\eta}_{1:T}$ is found through a brute-force grid search of threshold values $\eta_{t} \in [0,1]$ to minimize
\begin{equation}
\label{equation:t_float_loss_opt_2}
    \boldsymbol{\eta}^*_{1:T} = \underset{\boldsymbol{\eta}_{1:T}}{\textrm{argmin}} f\left(\Theta; \boldsymbol{x}_{1:T}\odot \left(\boldsymbol{m^*}_{1:T}>\boldsymbol{\eta}_{1:T} \right)
    \right) + \lambda_1^2 \norm{1 - \left(\boldsymbol{m^*}_{1:T}>\boldsymbol{\eta}_{1:T} \right)}_1,
\end{equation}
where $\lambda_1^2$ is a constant governing sparsity in the final mask $\boldsymbol{m^*}_{1:T}>\boldsymbol{\eta}_{1:T}$. Two simplifications make the brute-force optimization efficient while maintaining realistic representations. First, because the mask from the first optimization step is sparse and near-binary, the grid-search area can be limited to small regions around unique values found in $\boldsymbol{m^*}_{1:T}$. Second, the grid search can also be limited to optimizing \textit{backwards} through time, applying the same threshold across all features at a given time, heavily reducing grid search from $\boldsymbol{\eta}_{1:T} \in [0, 1]^{N \times T} \to \eta_{1:T} \in [0, 1]^T$. This also urges the mask to maintain clinical validity by ensuring that binarization of features at times $t \leq \tau$ does not affect the optimization for $t > \tau$. Finally, the binary mask is obtained by defining
\begin{equation}
\label{equation:m_final}
\boldsymbol{M^*}_{1:T} =\left(\boldsymbol{m}_{1:T}>\boldsymbol{\eta}_{1:T} \right).
\end{equation}

\subsubsection{Implementation Details}
An algorithmic description summarizing LBM's two-step process for generating a binary mask for an individual patient can be found in {\ref{section:appendix:b}}. The LBM was implemented by adding to the trained RNN model an additional layer that multiplies the input data with a trainable mask (first term of the cost function in Equation \ref{equation:m_loss_opt_1}). This enabled the leveraging of the same mechanics and infrastructure that were used to construct and train the RNN model. The trainable mask was initialized to all ones. Equation \ref{equation:m_float_loss_opt_2} was minimized using RMSProp \citep{tieleman2012lecture} with an initial learning rate of $0.1$. LBM hyperparameters were experimentally determined through trial and error and were chosen using qualitative examinations of the masks based on sparsity of the resulting mask and binarization of the first mask obtained from the first optimization step. The parameters were set as follows: $\lambda_1^1=0.005,\ \lambda_2^1=0.0005,\ A=-5$. The learning rate was reduced by a factor of 10 if the loss function did not improve after 5 iterations. Optimization was terminated when either 5000 iterations were reached or the training loss did not improve after reducing the learning rate 2 times. Equation \ref{equation:t_float_loss_opt_2} was optimized using brute-force grid search of threshold values backwards through time. 
Optimization was terminated when either each entry of the matrix $f\left(\Theta; \boldsymbol{x}_{1:T}\odot \boldsymbol{m}_{1:T} \right)$ was less than 0.05 or three iterations of searching back through time were reached.
Although there is no theoretical guarantee for the existence of a unique solution, our experiments indicate that the two-step approach was successful in finding a non-trivial binary mask that satisfies Equation \ref{equation:m_loss_opt_1} across all RNN patient predictions.
\subsection{KernelSHAP}
\label{section:methods_kshap}
Introduced in the 1950s, Shapley values answer a question from cooperative game theory: if $N$ players cooperated with each other for a collective reward, then how is the reward distributed fairly to each of the $N$ players {\citep{shapley1953value}}? Equivalently, what is the marginal contribution of each player? Consider the payout when a subset of $S$ players cooperate with each other without player $j$ and the payout when these same $S$ players cooperate with player $j$, then take the difference between these two payouts. This process is repeated over all possible $2^{N-1}$ subsets of players without player $j$, and the average of the resulting payout differences is the Shapley value for player $j$. Implicitly assumed in the formulation is the ability to observe the payout for each of the scenarios. The individual Shapley values sum to the collective reward.

This game theory principle was adapted to compute the ``payout'' of input features to a model's prediction, i.e. contributions of input variables. The $N$ players are the $N$ specific feature values at a particular input data instance that generated a specific prediction $y^*$, i.e. $f(x^*_1, \hdots, x^*_N) = y^*$, where $f$ denotes the model. Missing players correspond to input features whose values are unknown, and the payout in this scenario is the expectation of $f$ over all the possible values that these features could take. The required computations have exponential time complexity and render the method impractical. Approximation methods, including sampling techniques described in {\citep{vstrumbelj2014explaining}}, make the method tractable.

KernelSHAP incorporates the game theory principle of Shapley values with an existing model interpretation method, LIME (Local Interpretable Model-agnostic Explanation). The LIME framework finds a low-complexity function $g$ that approximates $f$ around a given point {\citep{ribeiro2016should}}. The inputs to $g$ are $\{z_j\}^m_{j=1}$, where $z^\prime_j$ is an interpretable combination of the input features $\{x^*_j\}^N_{j=1}$. For the purposes of this paper, $z^\prime_j=1$ if the value of feature $j$ is known, and $z^\prime_j=0$ otherwise. In other words, $z^\prime_j$ is a toggle for the presence or `missingness' of feature $j$; hence $m=N$. Let $h_x(z^\prime)$ denote the inverse mapping of $z^\prime(x)$. If $z^\prime_j=1$, then $(h_x)_j = x^*_j$; if $z^\prime_j=0$, then $(h_x)_j$ is unknown or random. For $g$ to locally approximate $f$, then $f(h_x(z^\prime)) = g(z^\prime)$ when $h_x(z^\prime)$ is near the data instance $x$. This means that if $g$ is linear, then in the neighborhood of a specific data instance, $x^*$, $f$ can be written as:
\begin{equation}
f(x^*) = \phi_0\ +\ \sum_{j=1}^N\phi_j\, z^\prime_j(x^*). \label{eqn:lime}
\end{equation}
Note that if none of the feature values are known (i.e. $z^\prime_j=0$ for all $j$), then $f=\phi_0$. This means that $\phi_0$ is the expected value of $f$ over the entire space, i.e. the ``background value.'' If all of the feature values are known ($z^\prime_j=1$ for all $j$), then $f(x^*)-\phi_0 = \sum_{j=1}^N \phi_j$. This says that the summation on the right is how much the model prediction changes from the background value if all $N$ input features take on the values in the data instance $x^*$. Lundberg and Lee showed that the coefficients $\phi_j$ ($j\geq 1$) are exactly the Shapley values if they solve the weighted least squares problem given by:
\begin{equation}
\underset{\phi\, \in\, R^{N+1}}{\textrm{argmin}}\ \sum_{z^\prime\, \in\, \{0,1\}^N} \pi_x(z^\prime) \left[f(h_x(z^\prime)) - \left(\phi_0 + \sum_{j=1}^N\phi_j z^\prime\right) \right]^2, \label{eqn:shap_lqs}
\end{equation}
where $\pi_x(z^\prime)$ is the Shapley Kernel {\citep{lundberg2017unified}}. The term $f(h_x(z^\prime))$ is the expected value of $f$ conditioned on $z^\prime$: if $z^\prime$ has ones in positions $i,j,k$ and zeros everywhere else, then $f(h_x(z^\prime))\ =\ E[f\,|\, x_i=x_i^*, x_j= x_j^*, x_k=x_k^*]$, where $x_i^*, x_j^*, x_k^*$ are the actual values of features $i,j,k$ in the specific input (data instance) that generated the prediction of interest.  One can think of Equation {\ref{eqn:shap_lqs}} as an over-determined system of linear equations relating the $\phi_j$'s with the expected values of $f$ when different combinations of features are unknown while the rest take their values from the data instance. The Shapley Kernel, $\pi_x(z^\prime)$, places higher weights to those equations corresponding to $z$ vectors with either a very small or very large number of ones (i.e. a very small or very large number of features inherit their values from the given data instance $x$). Since each element of $z^\prime$ is either 0 or 1, then the outer summation has $2^N$ terms, which is the number of all possible $z^\prime$ vectors. Further, for all but one of these vectors, $z^\prime$ has at least one zero element, and computing $f(h_x(z^\prime))$ involves taking the expectation of $f$ over the corresponding subspace of feature values. 

For practical use, implementations of Equation {\ref{eqn:shap_lqs}} must address two main problems: (1) efficiently estimate the expectation $f(h_x(z^\prime))$ for a given $z^\prime$; and (2) reduce the number of terms in the outer summation.
The first requires a background dataset from which to draw random values for $x_i$ when $z^\prime_i=0$. For some classes of $f$, e.g. tree-based models, fast versions for estimating the expectation have been proposed, e.g. Tree SHAP {\citep{lundberg2018consistent}}. KernelSHAP is the existing implementation of the general (model-agnostic) case, and in some of its applications, $f(h_x(z^\prime))$ is approximated by evaluating $f$ only once using a data vector where any `unknown' feature value (ie. where $z_j=0$) is set to what is considered a `normal' value for that feature, i.e. its median or mean {\citep{kshapgithub}}. Note that Equation {\ref{eqn:shap_lqs}} must be solved for each prediction that needs to be explained. In practice, if there are $T$ predictions, then the $T$ least squares problems are solved simultaneously. The total number of random $z^\prime$ samples for the collective outer summation is usually set to $2NT\ +\ 2048$ {\citep{kshapgithub}}. These samples are concentrated in regions where the the number of zeros in $z^\prime$ is very small or large since the kernel gives these regions higher weights.

Our implementation of KernelSHAP for the RNN mortality model computes the expectation $f(h_x(z^\prime))$ by setting $x^j_\tau=0$ when $z^\prime_j=0$ in the outer summation of Equation {\ref{eqn:shap_lqs}}. This single-point evaluation simplifies the computation of $f(h_x(z^\prime))$ and significantly reduces the run time. Recall that $x^j_\tau=0$ means the measurement for feature $j$ either did not deviate from the population mean (at the first timestep, $\tau=0$) or did not change from the previous timestep ($\tau\geq 1$). The latter typically resulted from having no new measurements in feature $j$; clinically, this meant the patient was considered stable {\citep{schulman2010standards}}. Therefore, our expectation for $f$ at a sample $z^\prime$ is equal to what the model would have predicted when the specified group of features (identified by zeros in $z^\prime$) had no new measurements. At the time of our implementation, the existing KerenelSHAP libraries did not properly compute Shapley values for multi-modal time-series data such as EMR, so additional modifications were made as described in {\citep{ferreira_2019}}.
\subsection{Interpreting RNN Predictions}
This section provides three different use cases demonstrating how the LBM and KernelSHAP can be used to determine the contributions of input features to the RNN model's ICU mortality predictions and identify which were most important at various scales or levels. These levels are 1) for an individual, 2) different subgroups defined by disease processes, and 3) the general ICU population. The common theme across the different levels is the repeated averaging of local information over specified time periods of a group of individuals. Section {\ref{section:results}} will illustrate with specific examples.

\subsubsection{Individual Attribution Matrices}
\label{section:methods:attribution_matrices}
Given the RNN model's sequence of mortality risk predictions for an individual \mbox{patient's} entire encounter, the LBM and KernelSHAP each generate an attribution matrix describing the contributions of input elements to those predictions. The attribution matrix, denoted by $\boldsymbol{a}^p_{1:T} \in \mathbb{R}^{N \times T}$ for patient encounter $p$, has the same dimensions as the encounter's dt-patient-matrix, $\boldsymbol{x}^p_{1:T}$.  
For retrospective analysis presented in this study, $T$ is the final hour before ICU discharge. Recall that $N$ is the number of input features at each timestep.

The LBM attribution matrix is given by $\boldsymbol{a}^p_{1:T} = 1 - \boldsymbol{M^*}_{1:T}$, where $\boldsymbol{M^*}_{1:T}$ is the mask defined by Equation \ref{equation:m_final}. Since $\boldsymbol{M^*}_{1:T}$ is binary, then $\boldsymbol{a}^p_{1:T}$ is also binary. For KernelSHAP, Equation {\ref{eqn:shap_lqs}} must be solved for each timestep $\tau \in [0,T]$, and the resulting coefficients $\phi_\tau^j$ comprise the elements of the KernelSHAP attribution matrix. The different formulations of the LBM and KernelSHAP mean that they generate different attribution matrices that can provide complementary perspectives.

\begin{itemize}
\item The LBM answers the question, ``Which of the non-zero elements of an individual's input dt-patient matrix led to the non-zero mortality predictions for this individual in $[0,T]$?'' The locations of ones in the LBM attribution matrix identify which of the non-zero entries in the dt-patient matrix were `zeroed' by the LBM to drive all the ROM predictions for patient encounter $p$ to zero. Equivalently, these non-zero changes in feature measurements from one timestep to the next provided evidence for the non-zero ROM predictions. For example, if $j$ corresponds to heart rate, and $a^j_\tau=1$, then the non-zero change in heart rate (e.g. decrease of 10 beats per minute) from time $\tau-1$ to $\tau$ led to the non-zero mortality predictions. The LBM attribution matrix does not describe how the prediction would change if the heart rate had increased by 30 bpm instead of decreased by 10 bpm. Neither does the LBM describe what would happen to the prediction if the heart rate had increased by 20 bpm instead of remained at the same value. It is important to note that the LBM will not highlight any input element $x_j^\tau$ that is already zero because setting the mask $m^j_\tau$ to zero for such elements does not change the value of $f$ in Equation {\ref{equation:m_float_loss_opt_2}} but increases the regularization term $\|1-\boldsymbol{m}_{1:T}\|$ in the loss function. Consequently, the LBM attribution matrix is sparse because the input dt-patient-matrix is sparse.

\item KernelSHAP expresses the prediction at time $\tau$ as a sum of contributions from the inputs at that time: \linebreak 
$y_\tau = \phi_0 + \sum_j \phi^j_\tau$.  
With $\phi_0$ denoting what the prediction would have been if no new measurements were recorded at that time, then the attribution matrix element $a^j_\tau \equiv \phi^j_\tau$ (which can be positive or negative) reflects how much would be added to $\phi_0$ given that the recorded change for feature $j$ at that time was in fact $x^j_\tau$. All the inputs at time $\tau$ cooperate with each other to generate $y_\tau$, and $a^j_\tau$ is the marginal contribution from knowing that the measurement for feature $j$ changed by exactly $x^j_\tau$ between $\tau-1$ and $\tau$. As a simple example, if the actual prediction at time $\tau$ was 0.45 and the background value was 0.15, then KernelSHAP expresses the difference, 0.30, as a sum of marginal contributions from the inputs: 0.25 was due to heart rate decreasing by exactly 10 bpm, -0.15 to systolic BP increasing by 30 mmHg, and 0.20 to the epinephrine dose remaining at 0.04 mcg/kg/minute from hour $\tau-1$ to hour $\tau$. Unlike the LBM, KernelSHAP's $a^j_\tau$ can be non-zero even when $x^j_\tau$ is zero. KernelSHAP does not answer the question: ``if $x^j_\tau$ had been different from its current value (regardless of what that current value is), then how would the prediction change?''
\end{itemize}

The attribution matrices generated by the LBM and KernelSHAP for a single patient encounter contain a lot of information. They can be aggregated in many different ways to facilitate different levels of analysis and enable  display of more concise information. Below we describe three aggregation techniques that each answer a different question:
\begin{itemize}
\item Which features were important during an individual patient's period of rapidly changing ROM predictions?
\item Which features were important to the RNN's ROM prediction in a specific diagnostic cohort of ICU patients \mbox{\textit{relative}} to the rest of the ICU population without that diagnosis? Equivalently, which features contribute to risk of mortality in a specific diagnostic group relative to the other ICU population?
\item Which features were important to the ROM predictions of the general population of critically ill children?
\end{itemize}

\subsubsection{Identifying Important Features for Individual Patient Predictions}
The goal is to identify which clinical features contributed to a critical period of illness of an individual patient, as indicated by rapidly changing ROM scores. Hence, we average the individual's attribution matrix over the time window of increasing ROM scores. In a deployment setting, a clinician could highlight any time window of interest to understand which features were contributing to the RNN ROM predictions. The averaging process reduces the amount of information from an $N\times T$ matrix to an $N$-dimensional vector of real values reflecting which features contributed, on average, to the ROM predictions \mbox{\textit{in that time window}}. 

Let $[t_i, t_f]$ denote a time interval of interest, e.g. where the ROM prediction for an individual patient increases significantly. The attribution matrix ${\boldsymbol{a}}^p_{1:T}$ can be regarded as a sequence of $N$-dimensional attribution vectors, ${\boldsymbol{a}}^p_\tau$, with $\tau$ denoting time. The vectors corresponding to the time points in $[t_i, t_f]$ are averaged as follows:
\begin{equation}
\label{equation: avg_timewindow}
\overline{\boldsymbol{a}}^p=\frac{1}{(t_f - t_i)}\sum_{\tau=t_i}^{t_f}| {\boldsymbol{a}}^p_\tau|, 
\end{equation}
where the absolute value is applied element-wise. For the LBM, the $j^{th}$ element of the resulting $N$-dimensional vector, $\overline{\boldsymbol{a}}^p$, reflects how often the $j^{th}$ input feature provided evidence for mortality within the specified time interval. For KernelSHAP, this element reflects how much feature $j$, on average during that time interval, changed the prediction (in magnitude) from the background value.

The vector $\overline{\boldsymbol{a}}^p$ can be normalized so that its largest element is unity, i.e. the feature that contributed the most is assigned a value of one:
\begin{equation}
\label{equation: avg_timewindow_b}
    \overline{\boldsymbol{a}}^p \leftarrow \frac{\overline{\boldsymbol{a}}^p}{\|\overline{\boldsymbol{a}}^p\|_\infty}.  
\end{equation}

Section {\ref{section:results:indiv}} illustrates with examples the computation of $\overline{\boldsymbol{a}}^p$ for two patient encounters from the test set. The two encounters were chosen to have different diagnoses but similar predicted ROM trajectories with a single inflection point, going from a low predicted ROM at ICU admission to a high predicted ROM near the end of the ICU stay. Two additional encounters are similarly analyzed in Appendix C, which were selected because their ROM predictions similarly had only one time period of substantial change and because they survived their ICU stay.

\subsubsection{Average Feature Attributions Within Cohorts: Relative Attribution Features} 
Instead of averaging only over specific time windows within individual attribution matrices, we can average over their entire ICU stay. This can further be done over a specific group of patients to understand the top contributing features for that group. As before, temporal averaging reduces each patient encounter's attribution matrix to a vector (parallel to Equation {\ref{equation: avg_timewindow}}). Next, these vectors are averaged over a group of encounters. When the group is defined by a disease process or diagnosis, this aggregation process is akin to traditional clinical research wherein logistic regression models are used to determine risk factors for that disease. For example, we can identify the features affecting mortality predictions in sepsis patients (here denoted by the set $S$) by computing the average absolute value of the attributions across patients whose primary diagnosis was sepsis:
\begin{equation}
\label{equation:risk_factors1}
\overline{\boldsymbol{a}}_S\ =\ \frac{1}{n(S)}\sum_{\forall p \in S}\,\frac{1}{t_p - t_0}\sum_{t=t_0}^{t_p}\, \abs{\boldsymbol{a}^p_\tau},
\end{equation}
where $t_0$ and $t_p$ are the first and last time points for patient $p$, and $n(S)$ is the number of patients in $S$. 
A similar computation was performed over all patients without a sepsis diagnosis, denoted by $S^c$, to yield $\overline{\boldsymbol{a}}_S^c$. Both $\overline{\boldsymbol{a}}_S$ and $\overline{\boldsymbol{a}}_S^c$ are $N$-dimensional vectors, where $N=398$ is the number of features used by the RNN model at each hour of prediction. While $\overline{\boldsymbol{a}}_S$ highlights which features were important in predictions for sepsis patients, $\overline{\boldsymbol{a}}_S^c$ identifies which features were pertinent in non-sepsis patient predictions. Normalizing them to unit-length vectors and then subtracting one from the other yields
\begin{equation}
\label{equation:risk_factors2}
    \tilde{\boldsymbol{a}}_S=\frac{\overline{\boldsymbol{a}}_S}{\norm{\overline{\boldsymbol{a}}_S}} - \frac{\overline{\boldsymbol{a}}_S^c}{\norm{\overline{\boldsymbol{a}}_S^c}}.
\end{equation}
We refer to the elements of the resulting $N$-dimensional vector, $\tilde{\boldsymbol{a}}_S$, as the \textit{relative attribution features} (RAFs) because they describe which features affect mortality prediction among sepsis patients more than they do non-sepsis ones. If the $j^{th}$ element of $\tilde{\boldsymbol{a}}_S$ is positive, then feature $j$ was more important in predicting mortality for sepsis patients than in non-sepsis patients. RAFs can be computed between two cohorts to determine which features affect mortality more in one cohort relative to the other. Two examples were explored: between patients with and without sepsis, and between patients with and without brain neoplasm. 

\subsubsection{Feature Contributions in All Critically Ill Children}
\label{section:methods:fimp}
We are also interested in the top contributing features for the general population including all critically ill children irrespective of diagnosis. For this, the aggregation in Equation {\ref{equation:risk_factors1}} is done over all patient encounters in the (test) set instead of only a subset of them, and the resulting vector normalized in a manner similar to Equation {\ref{equation: avg_timewindow_b}}:
\begin{equation}
\label{equation:overall_fimp}
    \overline{\boldsymbol{a}}_S \leftarrow \frac{\overline{\boldsymbol{a}}_S}{\|\overline{\boldsymbol{a}}_S\|_\infty}.  
\end{equation}
For the LBM, the $j^{th}$ element of $\bar a_S$ corresponds to the relative frequency that non-zero changes in feature $j$ (e.g. Heart Rate) led to non-zero ROM predictions in the general ICU population. It is ``relative'' in the sense that the most frequent feature is assigned a value of unity. For KernelSHAP, this value reflects the average contribution, in magnitude, of feature $j$ to the mortality probabilities for the population. In either case, $\bar a_S$ shows population-level feature contribution and can be considered as the ``weights'' of the RNN, similar to the weights of a logistic regression developed from the entire population. The vector $\bar a_S$ was computed using both the LBM and KernelSHAP attribution matrices of \textit{test} set encounters and is presented in Section \ref{section:results:fimp}. This computation of feature importance differs from standard \textit{permutation} feature importance computations which rank features based on their effects on model performance. Feature importance computations using the LBM and KernelSHAP attributions are derived from examining specific properties for each individual prediction; they are not optimized to improve model performance.

\subsection{Compute Time}
Lastly, to test the real-time clinical deployment potential of each method, the time-to-compute per patient was compared and is presented in Section \ref{section:results:speed}. Also included for comparison is the time-to-compute for generating RNN predictions. Because length of stay in the ICU varies greatly (and therefore the amount of time-steps in the patient data), it is expected that computing attribution matrices may also vary. Timings were done using a cuda-enabled NVIDIA Titan V GPU and python 2.7 with Keras and Tensorflow backend.

\section{Results}
\label{section:results}
\subsection{RNN Model Performance}
\label{section:results:rnn_perf}
\begin{figure}[!htbp]
\centering
\includegraphics[width=0.6\textwidth]{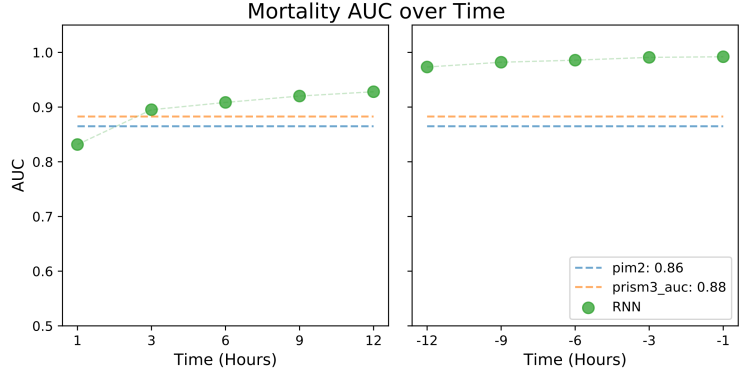}
\caption{Performance of the RNN model's predictions at significant times in the ICU}
\label{fig:rnn_auc}
\end{figure}

Figure \ref{fig:rnn_auc} shows the performance of the model at critical times in the ICU: the first twelve hours after admission and the last twelve hours before discharge (or death). At the $12^{th}$ hour after admission, the RNN ($0.93$) significantly outperformed PIM2 ($0.86$) and PRISM3-12 ($0.88$). From ICU admission to discharge, the RNN's AUC improved over time, approaching 1 at the end of stay. Over time as the RNN accrued more data and learned more about the patient and as lead time decreased {\citep{leisman2020development}}, the model's predictions became more accurate.

\subsection{Interpreting Individual Patient Predictions}
\label{section:results:indiv}
The LBM and KernelSHAP were applied to the RNN model's ROM predictions for all encounters in the test set. The attribution matrices are analyzed and presented here for two individual patients who subsequently died. These two patient encounters were selected for individual analysis  because their ROM predictions contained only one time window of substantial change from low to high mortality risk. Patients in the ICU can undergo multiple periods of volatility, as reflected in multiple shifts in the trajectory of ROM predictions, and limiting the patient-level analysis to encounters with only one such period simplifies the interpretation of results. See \ref{section:appendix:c} for analysis of two additional patients, which were selected similarly but with an additional criteria of surviving their ICU stay. The first patient, $p_1$, was an infant male weighing $6$ kg and diagnosed with pneumonia and acute respiratory distress syndrome (ARDS) {\citep{drw_ref_b}} caused by whooping cough (pertussis), severely affecting the respiratory system. The second patient, $p_2$, was a $3$ year old female weighing $27$ kg with a diagnosis of Grand mal status epilepticus {\citep{drw_ref_c}}, a condition of continuous seizures, caused by intracranial hemorrhage (bleeding in the cranium with elevated intracranial pressure) and ARDS.

\begin{figure*}[htbp!]
\centering
\includegraphics[width=1.0\textwidth]{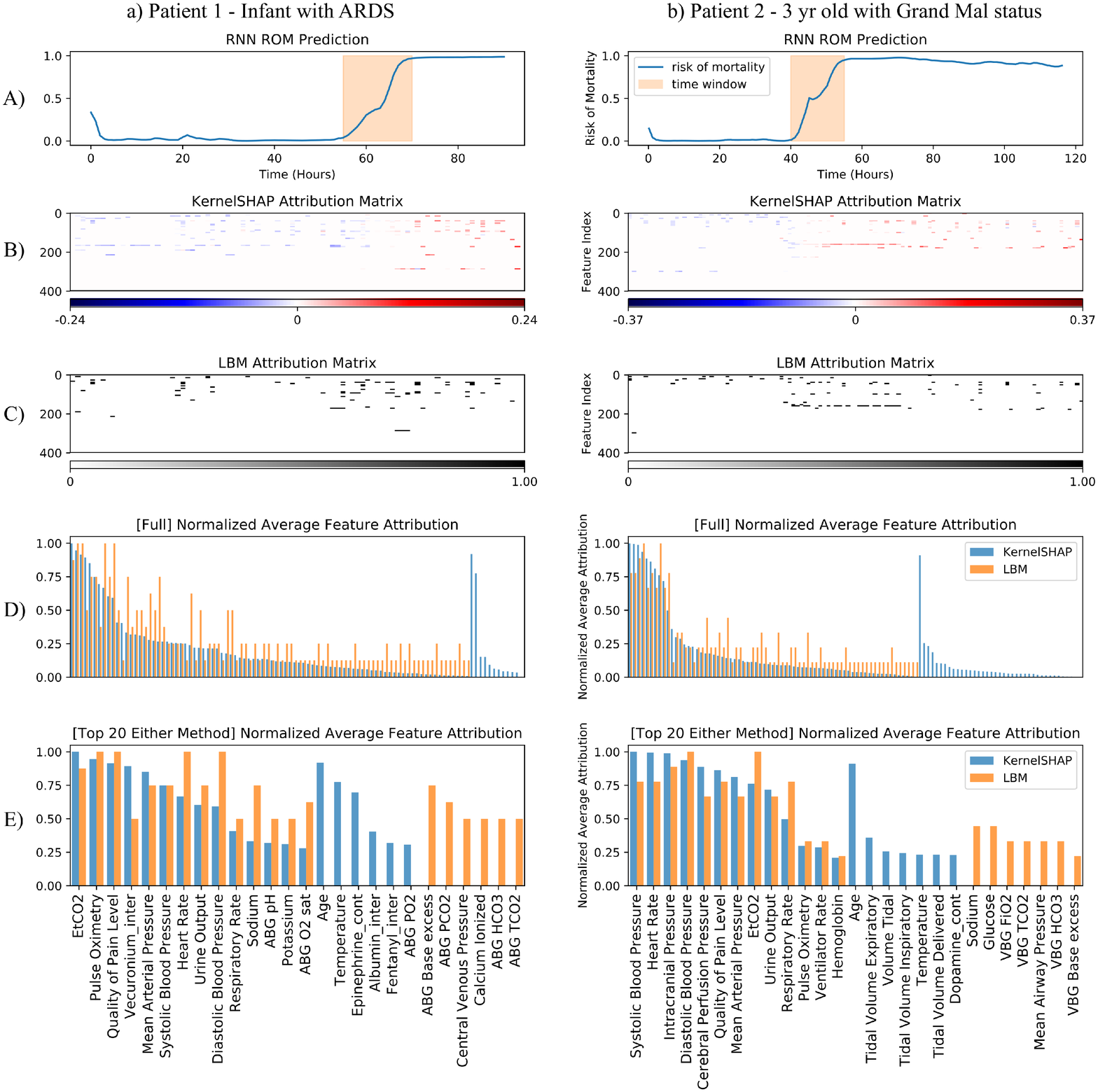}
\caption{Predictions and explanations for two individual encounters: a) $p_1$ and b) $p_2$. ROM predictions are visualized in A (top panel). KernelSHAP and LBM attribution matrices are shown as heatmaps in panels B and C, respectively. For KernelSHAP, the heatmap values range from negative to positive in probability units. For LBM, the heatmap is binary.
Attribution matrices were averaged over time periods highlighted in panel A using Equations {\ref{equation: avg_timewindow}} \& {\ref{equation: avg_timewindow_b}} to identify the features that contributed to the increasing ROM predictions in the highlighted time window. These features are visualized in panel D. Finally, panel E visualizes a subset of the features in panel D, presenting only the top 20 features from either method. Note that attribution matrices in B \& C are plotted with time on the x-axis (corresponding to ROM plots in A) and features on the y-axis. Also note that there could be more than 20 variables in E as the selected top 20 features overlap between the methods.}
\label{fig:indiv_plots}  
\end{figure*}

The RNN's ROM predictions are shown in Figure \ref{fig:indiv_plots}A-a for $p_1$ and \ref{fig:indiv_plots}A-b for $p_2$, while the attribution matrices are shown in Figure {\ref{fig:indiv_plots}}B (KernelSHAP) and {\ref{fig:indiv_plots}}C (LBM) as heatmaps, with time on the x-axis and features on the y-axis. In both patients, ROM started low but quickly increased over a $15$ hour window (shaded region in Figures \ref{fig:indiv_plots}A-a and \ref{fig:indiv_plots}A-b) to a high value. 
Equation {\ref{equation: avg_timewindow}} was used to identify which features were responsible for this ROM increase. In other words, the attribution matrices were averaged across the time periods of interest: $\overline{\boldsymbol{a}}^{p_1}  = \sum_{t=55}^{70} \boldsymbol{a}^{p_1} (t) / (70 - 55)$ and $\overline{\boldsymbol{a}}^{p_2} = \sum_{t=40}^{55} \boldsymbol{a}^{p_2} (t) / (55 - 40)$, then normalized according to Equation \ref{equation: avg_timewindow_b}. The results are presented as bar plots in Figure \ref{fig:indiv_plots}D, where features are ranked in descending order by KernelSHAP attributions. To simplify the amount of information for analysis, the top 20 features from either method are presented in Figure \ref{fig:indiv_plots}E, which makes it easy to see which features were common to both KernelSHAP and LBM (blue and orange bars together), which were identified only by KernelSHAP (blue bars only), and which were identified only by LBM (orange bars only).

Despite the similar increase in ROM predictions for these two patients, the features that contributed to their ROM trajectories differed. In the 15-hour window of interest, the increasing ROM predictions for $p_1$ were attributed to 177 (45\%) of the 398 input features, while the predictions for $p_2$ were attributed to 188 (47\%). Despite the model not having any input of specific diagnoses, the identified features are clinically consistent with each patient's diagnoses. The top features for the patient with pneumonia and ARDS ($p_1$) are either related to the respiratory system (EtCO2, Pulse Oximetry, Respiratory Rate) {\citep{drw_ref_b}} or associated with infections (Temperature, Heart Rate, Systolic/Diastolic/Mean Arterial Blood Pressure, and 5 blood gas measurements (ABG pH, ABG PO2, ABG PCO2, ABG HCO3, ABG TCO2) {\citep{drw_ref_e}}. Two of the top ranked features for the patient with seizures ($p_2$) -- Intracranial Pressure (rank 3) and Cerebral Perfusion Pressure (rank 5) -- align with her diagnoses of intracranial hemorrhage. In addition, contributing features for this individual's increase in predicted ROM included EtCO2, Pulse Oximetry, and Respiratory Rate (rank 8, 11, and 10 respectively), which align with her diagnosis of ARDS {\citep{drw_ref_b}}. Other top contributing features for this individual -- Heart Rate, Systolic/Diastolic/Mean Arterial Blood Pressure -- are general markers of critical illness 
\mbox{\citep{nichols2016rogers, pollack1996prism, slater2003pim2}}.

\begin{figure*}[!htbp]
\centering
\includegraphics[width=1.0\textwidth]{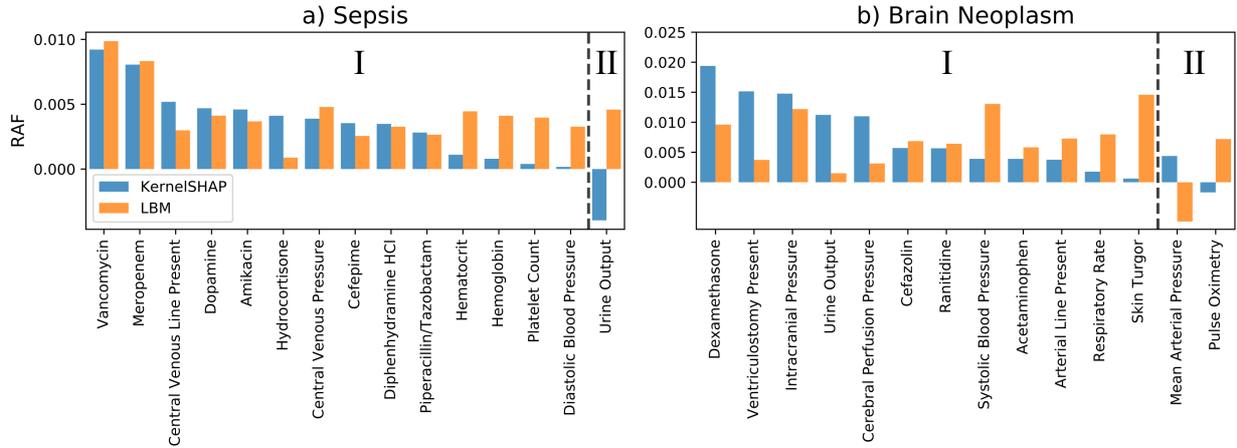}
\caption{Top 10 Relative attribution features (RAFs) computed across two primary diagnoses: a) sepsis, b) brain neoplasm. Further, the plots are partitioned into two regions: I) when KernelSHAP and LBM align; and II) when KernelSHAP and LBM disagree. Note that there could be more than 10 variables as the top 10 features overlap between the methods.}
\label{fig:cohort}
\end{figure*}

\subsection{Average Feature Attributions Within Cohorts (Relative Attribution Features)}
\label{section:results:cohorts}
The attribution matrices from KernelSHAP and LBM were aggregated using Equations \ref{equation:risk_factors1} \& \ref{equation:risk_factors2} to compute
relative attribution features (RAF) across two sub-populations of primary disease diagnoses: a) sepsis $(n(S)=122)$, b) brain neoplasm $(n(S)=112$); recall n(S) is the number of patients in sub-population $S$. The features with the top 10 RAF from either method are shown in Figure \ref{fig:cohort}. Each bar plot is further partitioned into two regions: I) when both KernelSHAP and LBM align (i.e. both are positive); and II) when KernelSHAP and LBM disagree. Features in Region I are variables which both methods identified as affecting predicted ROM to a greater extent in the specific disease cohort relative to the general critically ill population. Note that features with negative RAF (denoting higher importance in the general critically ill population cohort, $S^{c}$, than in the specified disease cohort) from both methods were not included. Features in Region II are those where the methods disagreed. The following paragraphs provide a general description of each disease as well as relevant observations from Figure \ref{fig:cohort}.

\paragraph{Sepsis}
Sepsis is a condition in which the body responds to a severe infection by releasing chemicals which can also damage multiple organ systems, cause hemodynamic instability, and result in abnormal blood counts. Sepsis is treated by treating the underlying infection with antibiotics and maintaining blood flow, fluids, and organ function {\citep{drw_ref_e}}. Not surprisingly, therefore, five of the features in Region I are antibiotics (Vancomycin, Meropenem, Amikacin, Cefepime, Piperacillin/Tazobactam). Another three are measures of blood counts (Hematocrit, Hemoglobin, and Platelet Count), three are related to blood pressure measurements (Central Venous Pressure/Line Present, Diastolic Blood Pressure), one is a drug to increase blood pressure (Dopamine), and another is a drug often used in sepsis to treat hemodynamic instability (Hydrocortisone).

\paragraph{Brain neoplasm} Brain neoplasms cause increased intracranial pressure and altered neurologic function {\citep{drw_ref_f}}. Treatments include maintaining normal intracranial pressure using interventions such as drugs and shunts. Of the twelve variables in Region I, two are related to measurements of brain pressure and perfusion (Intracranial Pressure, Cerebral Perfusion Pressure), and three are interventions associated with elevated intracranial pressure (Ventriculostomy, Dexamethsaone, Ranitidine).

\subsection{Feature Contributions in All Critically Ill Children}
\label{section:results:fimp}
\begin{figure*}[htbp!]
\centering
\includegraphics[width=.9\textwidth]{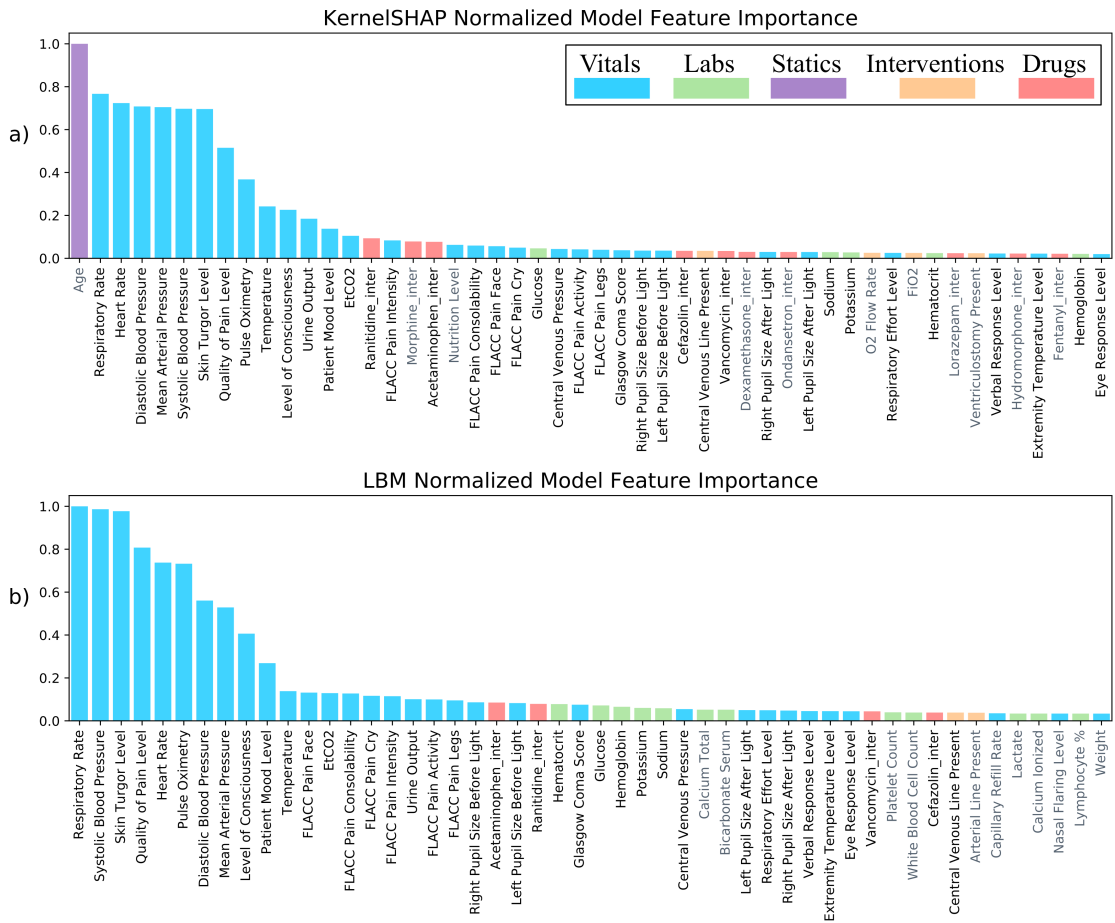}
\caption{Top 50 features from model ``feature importance'' computed using a) KernelSHAP and b) LBM. Variables are colored by their variable types, denoted in the legend. Variables common to both KernelSHAP's top 50 and LBM's top 50 are indicated with darker text labels.}
\label{fig:fimp}  
\end{figure*}

Using Equations  {\ref{equation:risk_factors1}} \& {\ref{equation:overall_fimp}}, KernelSHAP and LBM attribution matrices were averaged for \textit{all} critically ill children at all times in the test set to compute a form of model feature importance. This population consisted of 2008 patient encounters. Because the importance from both methods decays rapidly, only the top 50 are investigated. The top 50 features from both methods are displayed in Figure \ref{fig:fimp}. Features are colored by their variable type: vitals in blue, labs in green, interventions in orange, drugs in red, and statics (demographics data) in purple. Variables that were common to both KernelSHAP's and LBM's top 50 are indicated with darker text labels while those that are not are labeled in grey. 

Because different mortality models often use different feature sets, it is difficult to compare the feature importances computed here for the RNN model with weights from other models such as PRISM3 or PIM2, which often use variables measured specifically for the algorithm and not recorded in the EMR. A notable observation from Figure \ref{fig:fimp} is that a majority of the top features extracted by both KernelSHAP and LBM were variables measured \textit{internally} to the patient, i.e. vitals, labs, age, and gender. This is enumerated in Table \ref{tbl:fimp_vtype_perc} and is consistent with previously reported results \citep{ho2017dependence}, which showed little to no degradation in model AUC when external variables (interventions and drugs) were excluded from the input patient representation. In particular, such fundamental measures of the status of critically ill children such as heart rate, blood pressures, blood gases, pulse oximetry (which are routinely monitored in critically ill children), skin turgor, and observations reflecting the level of consciousness (pain perception, comfort, etc) \mbox{\citep{nichols2016rogers, pollack1996prism, slater2003pim2}} rank so highly. These observations give us confidence that the RNN model reasonably used the input features to generate risk of mortality predictions, and that the LBM and KernelSHAP extracted this information.

\begin{table}[!htbp]
\centering
\caption{Count (and percentage) of each variable type in the top 50 feature importance from KernelSHAP and LBM in Figure \ref{fig:fimp}.}
\label{tbl:fimp_vtype_perc}
\begin{tabular}{@{}l|ll@{}}
\toprule
Variable Type    & KernelSHAP & LBM       \\ \midrule
Vitals           & 30 (60\%)  & 32 (64\%) \\
Labs             & 5 (10\%)   & 12 (23\%) \\
Interventions    & 4 (8\%)    & 2 (4\%)   \\
Drugs            & 10 (20\%)  & 4 (8\%)   \\
Statics (Others) & 1 (2\%)    & 0 (0\%)   \\ \bottomrule
\end{tabular}
\end{table}

\subsection{Compute Time}
\label{section:results:speed}
\begin{figure}[!htbp]
\centering
\includegraphics[width=0.6\textwidth]{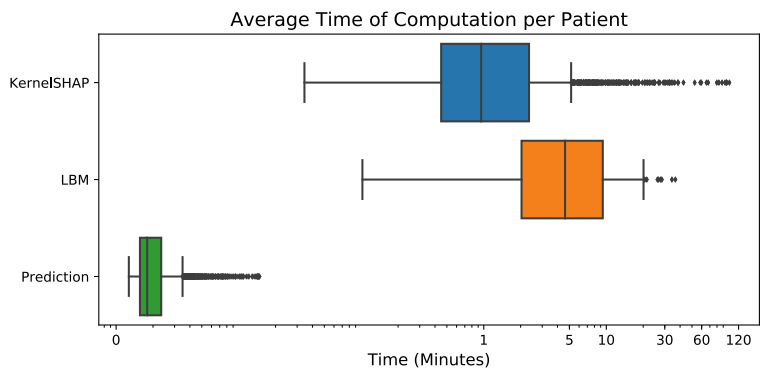}
\caption{Box-and-whiskers plot showing distribution of time-to-compute per patient for KernelSHAP and LBM. For reference, prediction time-to-compute per patient is also included.}
\label{fig:speed}
\end{figure}

Figure \ref{fig:speed} shows the average computational time of each method across all patients in the validation and test set. The median (IQR) time to compute predictions for a patient is 0.002 (0.001) minutes. To interpret the same predictions, KernelSHAP takes a median (IQR) time of 0.95 (1.90) minutes and LBM takes 4.60 (7.33) minutes. It should be noted that KernelSHAP can be much more computationally expensive for large feature spaces (398 here), however, the \textit{dt-patient-matrix} representation enables practical use of KernelSHAP to execute in real-time.

\section{Discussion}
\label{section:discussion}
Methods for interpreting deep learning models are very diverse in method and purpose. This study focused on determining the contribution of model inputs to model predictions. As a proof of concept, we have described in detail two methods that provide information about which clinical features made the most important contributions to risk of mortality predictions generated by a previously well described recurrent neural network (RNN) model using electronic health data of critically ill children. The first, Learned Binary Mask (LBM), is a newly described novel method based on occlusion principles to determine which inputs led to the mortality predictions of the many-to-many RNN ICU mortality model using multi-model EMR data. The second entailed modifying a previously described and familiar methodology, KernelSHAP, to perform on the same RNN model. While methods for interpreting deep learning models exist, most are not compatible with \mbox{many-to-many} RNN models whose inputs are multi-modal EMR data that include time series measurements. This motivated the approaches we took in this study with the LBM and KernelSHAP.

We focused this initial study of the LBM and KernelSHAP on the RNN mortality model. Given this model, its inputs, and the resulting trajectory of risk of mortality (ROM) predictions for an individual patient encounter, the LBM and KernelSHAP each estimate an attribution matrix that reflects each input's contribution to those predictions. Because the LBM and KernelSHAP have different formulations, they generate different attribution matrices for the same patient encounter. Each matrix answers different questions. Having both perspectives affords a more comprehensive picture of the patient than having either one alone.   

The analyses of individual patients in Section {\ref{section:results:indiv}} and \mbox{\ref{section:appendix:c}} show that the RNN model used patient-specific features, in addition to general markers for critical illness, for generating ROM predictions, highlighting the individual nature of the RNN predictions and what led to them.  Aggregating individual attributions across disease cohorts and the entire population of critically ill children enabled demonstrations of the LBM and KernelSHAP, as well as the RNN model, on a larger scale than analyzing individual patients. The evaluations were all qualitative: features identified by the LBM and KernelSHAP as important contributors to the ROM predictions for specified cohorts (sepsis patients, brain neoplasm patients, general ICU) were consistent with generally described clinical characteristics of their respective disease processes \mbox{\citep{drw_ref_e,drw_ref_f}}.  The results (Section {\ref{section:results:cohorts}}, Section {\ref{section:results:fimp}}) support the notion that the LBM and KernelSHAP identify clinical features most relevant in an individual patient, a specific disease cohort and the entire critically ill population.

Some interesting observations arose from analyzing these feature attributions. Not surprisingly, vital signs dominated the top contributing features for the entire population (Figure {\ref{fig:fimp}}) because they are general markers of critical illness. While a few drugs and interventions were also in the list of top 50 contributors, their contributions as measured by LBM or KernelSHAP were small compared to those of the vital signs. In contrast, several of the important contributors to individual patient ROM scores included drugs and interventions (Figure {\ref{fig:indiv_plots}}, Figure {\ref{fig:indiv_plots_appendix}}). These contrasting observations likely result from the way the attribution matrices were aggregated and averaged. The population-level aggregation (Figure {\ref{fig:fimp}}) included the entire ICU stay of each patient, while the patient-level aggregation (Figure {\ref{fig:indiv_plots}}) was confined to a limited time window of the \mbox{patient’s} ICU stay, selected for instability: where the ROM predictions were rapidly increasing. The window of increasing ROM may be regarded as a period of volatility for a patient, and during this period, drugs and interventions were important. The contributions of these drugs were diluted during the averaging process over the entire ICU stay of a patient who was stable for a majority of that stay. The difference in importance of drugs and interventions between Figure {\ref{fig:indiv_plots}} and Figure {\ref{fig:fimp}}, therefore, likely results from the fact that most patients survived their ICU encounter (96\%) and spent the majority of their ICU stay in a \mbox{``stable’’} \mbox{(non-volatile)} state.

The analyses of attributions in disease-specific cohorts (Figure {\ref{fig:cohort}}) also had noteworthy findings. In the sepsis group, 13 out of the 15 features with the highest RAF were consistent with clinical expectations {\citep{drw_ref_e}}; in the brain neoplasm group, it was 5 out of the top 11 {\citep{drw_ref_f}}. This is significant because patients often have multiple diseases, yet the aggregated results identified features that were consistent with clinical knowledge on these cohorts' diagnoses. We note, however, that this general alignment potentially results from the inherent bias in EMR data. For example, antibiotics are more often given to sepsis patients, and the computed RAF may simply reflect that bias. One possible approach to distinguish between data biases and actual impact on mortality predictions would be to compare the RAFs with the frequency that the variables were measured or administered; this is left for future work. 

The LBM and KernelSHAP identified features that are not specifically associated with the specified diagnoses of the individuals or the disease cohorts. It is important to note that these apparently non-specific features actually reflect critical illness and are in fact features common to many severity of illness scores \mbox{\citep{pollack1996prism, slater2003pim2}}. The LBM and KernelSHAP were formulated to determine features that contributed to patients' risk of ICU mortality, which reflect their severity of critical illness; they were not formulated to identify risk factors for specific diagnoses. Therefore, it is not surprising -- in fact, it is reassuring -- that they identified general markers of critical illness. It is remarkable that they also identified features specific to a patient's diagnoses despite having no information about those diagnoses.

Figure {\ref{fig:fimp}} shows that of the top 50 contributing features separately identified by the LBM and KernelSHAP, 39 (78\%) were shared, illustrating remarkable consilience between the two methods. Consilience between the two methods was also seen in the analyses of individual patients (Section {\ref{section:results:indiv}} and \mbox{\ref{section:appendix:c}}) and disease-cohorts (Figure {\ref{fig:cohort}}). Some concordance between the results from the two methods is not surprising and is, in fact, reassuring because they are conceptually similar in goal: to identify the set of features that contributed to the current prediction.

Nevertheless, the two methods formulate feature contribution differently. LBM identifies contributing features by finding the inputs that must be \mbox{\textit{zeroed}} to drive the mortality predictions to zero (i.e. provide evidence for non-zero mortality predictions). KernelSHAP expresses a prediction as a sum of a background value and marginal contributions from the inputs. The different perspectives and objective functions of the LBM and KernelSHAP gave rise to the differences between their attribution matrices. In particular, they treat zero-valued or near zero-valued inputs differently. LBM does not mask inputs that are exactly zero-valued because masking such inputs adds no benefit but increases the objective function the LBM is minimizing. Masking inputs that are near zero also incurs a high cost in the LBM objective function; therefore, there is a very small chance of LBM selecting inputs that are near zero. In contrast, KernelSHAP has no such restrictions against zero-valued or near zero-valued inputs. The hour-to-hour changes of most drugs and interventions are zero for extended periods. Consequently, these features were not significant contributors in the LBM formulation, but were in the KernelSHAP formulation. Age is also treated differently by the two methods, as was seen in the individual case studies (Figure 4E, Figure C.8E) and in the general ICU population (Figure 6). All entries for age after the initial timestep, are exactly an hour, which is negligible in terms of years, the unit used to express age. In the initial timestep, the entry for age is the deviation of the \mbox{patient’s} age from the \mbox{population’s} mean age, and this varies across different patients. While the LBM can mask age at the initial time step, it very likely ignores the negligible 1-hour age change at all subsequent timesteps. Consequently, the aggregation over many timesteps dilutes any contribution that may come from the initial timestep. Mechanisms related to KernelSHAP alone, such as the ad hoc approximations to make the Shapley equation (Equation 8) computationally tractable, likely gave rise to some of the other differences observed between the two methods' attribution matrices and the resulting aggregations.

While the results of applying the LBM and KernelSHAP to the RNN mortality model are promising, some questions remain. The LBM and KernelSHAP minimize objective functions that may have multiple local minima, and the actual numerical approximations of these minima depend on hyperparameters. The stability of these solutions, i.e. how the resulting feature contributions change with these solutions, has yet to be investigated. Understanding the sensitivity of the resulting clinical interpretations with respect to perturbations caused by artificial numerical issues is important. Closely related questions are the stability of the LBM or KernelSHAP outputs with respect to the RNN model itself, and with respect to the model inputs. For LBM, even a very slight change in an \mbox{input’s} value could flip an attribution matrix element from 0 to 1 (or 1 to 0) because of the hard thresholding (Equation {\ref{equation:m_final}}) to make the attributions binary. The aggregation of these elements over many time points can smooth this behavior. 

Exploring generalizability of the LBM is another avenue for future work. This will mean applying the LBM to different clinical datasets, tasks, and models. Finally, the aggregation examples presented here -- averaging over a narrow time period of an individual patient, averaging over the entire ICU stay of a patient then aggregating over many encounters -- focused on retrospective use and represent only a small sampling of possibilities. Perhaps the most promising use for clinical deployment is demonstrated by  time windows in individual patients. This would allow identifying periods of instability when the ROM is rapidly changing and determining what features make the major contribution to these changes. This approach could potentially direct attention to underlying contributions to this instability and direct therapeutic interventions. What and how to present information from the individual attribution matrices computed by the LBM or KernelSHAP will depend on the actual use case. Deploying the RNN model with the LBM and KernelSHAP for real-time clinical use has many considerations involving many different areas of understanding \mbox{\citep{kitzmiller2019diffusing, keim2018advancing}}, and these will require their own separate investigations. 

We emphasize that this study is not about feature selection for model development. Feature selection methods typically are used to prune variables from a large subset of potential input features to improve model performance as measured by metrics such as area under the curve (for classification tasks) or mean absolute errors (for regression tasks). Although the LBM and KernelSHAP could potentially be used for this purpose, it was not the purpose here. Instead, they were  used to determine and understand the contributions of predefined inputs to the predictions that have already been made by an existing (already trained) model.

\section{Conclusion}
\label{section:conclusion}
This proof-of-concept study presented two methods for providing information about which features made the most important contributions to the risk of mortality predictions of a previously described RNN model using the EMR of critically ill children. The first, Learned Binary Mask (LBM), is a novel occlusion-based method developed here. The second is an existing explanation method, KernelSHAP, modified in this study to make it compatible with the same RNN model. A novel representation of patient data enabled practical application of both methods on the many-to-many RNN model. For any given patient, each method generated an attribution matrix that reflects how each input contributed to the \mbox{RNN’s} predictions for that patient. The individual contributions were  aggregated across different scales - a specified time window of an individual, or entire encounters of disease subgroups or the whole population of critically ill children in the ICU -- to determine the clinical features which were most important to the mortality risk predictions for the specified cohort. Because the two methods have different formulations, their outputs provide complementary perspectives. While initial results show promise -- some consistency with established clinical knowledge -- several important questions about both the LBM and KernelSHAP remain open for investigation. These include, but are not limited to, more comprehensive clinical evaluations of their results; the stability of their solutions; and their generalization to other datasets, tasks, and models.

\section*{Funding}
This work was supported by the L. K. Whittier Foundation.

\bibliographystyle{elsarticle-harv}
\bibliography{bibliography.bib}

\pagebreak
\appendix
\section{Datasets Summary \& Feature List}
\label{section:appendix:a}
\floatsetup[table]{capposition=top}
\begin{table}[htbp!]
    \label{table:dataset_summary}
    \centering
    \caption{Summary statistics and demographics, partitioned into training (60\%), validation (20\%), and test (20\%) sets. Presented are the total number of encounters per category with percentages rounded to the nearest integer in parenthesis.}
\begin{tabular}{@{}r|lll@{}}
\toprule
                                         & \multicolumn{1}{c}{\textbf{Train N (\%)}} & \multicolumn{1}{c}{\textbf{Valid N (\%)}} & \multicolumn{1}{c}{\textbf{Test N (\%)}} \\ \midrule
\multicolumn{1}{l|}{\textbf{Mortality}}  & \multicolumn{1}{c}{}                      & \multicolumn{1}{c}{}                      & \multicolumn{1}{c}{}                     \\
Survived                                 & \multicolumn{1}{c}{5645 (96)}             & \multicolumn{1}{c}{1885 (96)}             & \multicolumn{1}{c}{1925 (96)}            \\
Died                                     & \multicolumn{1}{c}{240 (4)}               & \multicolumn{1}{c}{77 (4)}                & \multicolumn{1}{c}{83 (4)}               \\ \midrule
\multicolumn{1}{l|}{\textbf{Age Group}}  & \multicolumn{1}{c}{}                      & \multicolumn{1}{c}{}                      & \multicolumn{1}{c}{}                     \\
{[}0, 1)                                 & \multicolumn{1}{c}{976 (17)}              & \multicolumn{1}{c}{371 (19)}              & \multicolumn{1}{c}{309 (15)}             \\
{[}1, 5)                                 & \multicolumn{1}{c}{1449 (25)}             & \multicolumn{1}{c}{481 (25)}              & \multicolumn{1}{c}{496 (25)}             \\
{[}5, 10)                                & \multicolumn{1}{c}{1052 (18)}             & \multicolumn{1}{c}{331 (17)}              & \multicolumn{1}{c}{362 (18)}             \\
{[}10, 18)                               & \multicolumn{1}{c}{1941 (33)}             & \multicolumn{1}{c}{644 (33)}              & \multicolumn{1}{c}{683 (34)}             \\
18+                                      & \multicolumn{1}{c}{467 (8)}               & \multicolumn{1}{c}{135 (7)}               & \multicolumn{1}{c}{158 (8)}              \\ \midrule
\multicolumn{1}{l|}{\textbf{Primary Diagnosis Category}} &                                           &                                           &                                          \\
Respiratory                              & 1682 (29)                                 & 579 (30)                                  & 558 (28)                                 \\
Neurologic                               & 884 (15)                                  & 232 (12)                                  & 289 (14)                                 \\
Oncologic                                & 670 (11)                                  & 214 (11)                                  & 233 (12)                                 \\
Injury/Poisoning/Adverse Effects         & 583 (10)                                  & 228 (12)                                  & 185 (9)                                  \\
Orthopedic                               & 448 (8)                                   & 158 (8)                                   & 153 (8)                                  \\
Infectious                               & 417 (7)                                   & 155 (8)                                   & 153 (8)                                  \\
Other                                    & 339 (6)                                   & 125 (6)                                   & 118 (6)                                  \\
Gastrointestinal                         & 288 (5)                                   & 91 (5)                                    & 115 (6)                                  \\
Genetic                                  & 243 (4)                                   & 80 (4)                                    & 73 (4)                                   \\
Cardiovascular                           & 189 (3)                                   & 56 (3)                                    & 81 (4)                                   \\
Renal/Genitourinary                      & 142 (2)                                   & 44 (2)                                    & 50 (2)                                   \\ \bottomrule
\end{tabular}
\end{table}

\floatsetup[table]{capposition=top}
\begin{table*}[htbp!]
    \label{table:feature_list}
    \centering
    \caption{List of 398 features used as inputs to the RNN model}
    \resizebox{\textwidth}{!}{
    \begin{tabular}{lllll}

Pulse Oximetry & Heart Rate & Respiratory Rate & Weight & Systolic Blood Pressure\\Diastolic Blood Pressure & Mean Arterial Pressure & Motor Response Level & Verbal Response Level & Eye Response Level\\Glascow Coma Score & Temperature & Right Pupillary Response Level & Level of Consciousness & Left Pupillary Response Level\\Extremity Temperature Level & Patient Mood Level & Respiratory Effort Level & Capillary Refill Rate & Skin Turgor turgor\\Right Pupil Size Before Light & Left Pupil Size Before Light & Nasal Flaring Level & Right Pupil Size After Light & Left Pupil Size After Light\\Quality of Pain Level & Height & FLACC Pain Face & FLACC Pain Legs & FLACC Pain Activity\\FLACC Pain Cry & FLACC Pain Consolability & FLACC Pain Intensity & Nutrition Level & Lip Moisture Level\\Capillary Refill Delayed & Age & Sex F & Foley Catheter Volume & Sodium\\Potassium & Glucose & Hematocrit & Hemoglobin & Creatinine\\Bicarbonate Serum & Central Venous Pressure & Head Circumference & PaO2 to FiO2 & Chloride\\Calcium Total & BUN & Platelet Count & White Blood Cell Count & RBC Blood\\MCH & MCV & MCHC & RDW & O2 Flow Rate\\Lymphocyte Percent & Neutrophils Percent & Monocytes Percent & Basophils Percent & Oxygenation Index\\Eosinophils Percent & Calcium Ionized & Lactate & Albumin Level & ALT\\AST & Bilirubin Total & Alkaline phosphatase & Protein Total & PTT\\INR & Abdominal Girth & PT & CBG PCO2 & CBG pH\\CBG PO2 & CBG HCO3 & CBG Base excess & CBG TCO2 & CBG O2 sat\\Phosphorus level & Magnesium Level & Bands Percent & ABG PO2 & ABG pH\\ABG PCO2 & ABG O2 sat & ABG HCO3 & ABG Base excess & ABG TCO2\\VBG HCO3 & VBG Base excess & VBG TCO2 & VBG PO2 & VBG pH\\VBG PCO2 & VBG O2 sat & Culture Blood & C-Reactive Protein & Fibrinogen\\CBG FiO2 & VBG FiO2 & ABG FiO2 & Culture Urine & Influenza Lab\\Schistocytes & Metamyelocytes Percent & Culture Respiratory & Myelocytes Percent & Triglycerides\\Lipase & Culture CSF & MVBG Base Excess & MVBG HCO3 & MVBG PCO2\\MVBG PO2 & MVBG TCO2 & MVBG pH & MVBG O2 Sat & MVBG FiO2\\Oxygen Mode Level & FiO2 & Central Venous Line Site & EtCO2 & PEEP\\Peak Inspiratory Pressure & Ventilator Rate & Inspiratory Time & Mean Airway Pressure & Chest X Ray\\Arterial Line Site & Pressure Support & Tidal Volume Expiratory & Tidal Volume Inspiratory & Tidal Volume Delivered\\Volume Tidal & CT Brain & EPAP & MRI Brain & NIV Set Rate\\Ventriculostomy Site & Chest Tube Site & Abdominal X Ray & Hemofiltration Therapy Mode & ECMO Hours\\Acetaminophen inter & Ranitidine inter & Gastrostomy Tube Volume & Morphine inter & Lorazepam inter\\Ondansetron inter & Vancomycin inter & Fentanyl inter & Furosemide inter & Intracranial Pressure\\Cefazolin inter & Diphenhydramine HCl inter & Pantoprazole inter & Fentanyl cont & Dexamethasone inter\\Midazolam HCl inter & IPAP & Potassium Chloride inter & Ceftriaxone inter & Budesonide inter\\Piperacillin/Tazobactam inter & Dopamine cont & Dexmedetomidine cont & Hydromorphone inter & Vecuronium inter\\CSF RBC & CSF WBC & Methylprednisolone inter & Levetiracetam inter & Midazolam HCl cont\\Bilirubin Conjugated & Bilirubin Unconjugated & D-dimer & Macrocytes & Ibuprofen inter\\Diazepam inter & Alteplase inter & Ketorolac inter & Amylase & Rocuronium inter\\Culture Fungus Blood & Ceftazidime inter & Spherocytes & Meropenem inter & Sodium Chloride inter\\Famotidine inter & Cerebral Perfusion Pressure & Albuterol inter & Sodium Bicarbonate inter & Calcium Chloride inter\\Albumin inter & Magnesium Sulfate inter & CSF Lymphs Percent & Oxycodone inter & Clindamycin inter\\Metronidazole inter & Culture Wound & Phenobarbital inter & Fluconazole inter & Chlorothiazide inter\\Lansoprazole inter & Glycopyrrolate inter & CSF Glucose & CSF Protein & Azithromycin inter\\Potassium Phosphate inter & Atropine inter & Propofol inter & TSH & Reticulocyte Count\\Trimethoprim/Sulfamethoxazole inter & CSF Segs Percent & Racemic Epi inter & Lactate Dehydrogenase Blood & Hydrocortisone inter\\Acyclovir inter & Nifedipine inter & T4 Free & Baclofen inter & Acetaminophen/Hydrocodone inter\\Cefotaxime inter & Methadone inter & Ampicillin/Sulbactam inter & Ferritin Level & Acetaminophen/Codeine inter\\GGT & Tacrolimus inter & Ketamine inter & Nystatin inter & Gabapentin inter\\Micafungin inter & ESR & Clonidine HCl inter & B-type Natriuretic Peptide & Ferrous Sulfate inter\\Tobramycin inter & Prednisone inter & Enalapril inter & Amikacin inter & Oseltamivir inter\\Desmopressin inter & Insulin inter & Vitamin K inter & Naloxone HCL cont & Lactobacillus inter\\Sodium Phosphate inter & Calcium Gluconate inter & Propofol cont & Ampicillin inter & Fluticasone inter\\Olanzapine inter & Spironolactone inter & Aspirin inter & Isradipine inter & Acetazolamide inter\\Metoclopramide inter & Amlodipine inter & Montelukast Sodium inter & Amphotericin B Lipid Complex inter & Immune Globulin inter\\Heparin inter & Cefepime inter & Levocarnitine inter & Gentamicin inter & Lidocaine inter\\Topiramate inter & Filgrastim inter & Labetalol inter & Ursodiol inter & Fosphenytoin inter\\Voriconazole inter & CT Chest & Mycophenolate Mofetl inter & Valproic Acid inter & Anti-Xa Heparin\\Clonazepam inter & Epinephrine cont & Levalbuterol inter & Sucralfate inter & Aminophylline cont\\Oxcarbazepine inter & Ciprofloxacin HCL inter & Ipratropium Bromide inter & Furosemide cont & Levothyroxine Sodium inter\\Hydromorphone cont & Insulin cont & Vasopressin cont & Heparin cont & Epoetin inter\\Nitric Oxide & Blasts Percent & Amoxicillin inter & Epinephrine inter & Cephalexin inter\\Lactic Acid Blood & Aminophylline inter & Sildenafil inter & Enoxaparin inter & Chloral Hydrate inter\\Risperidone inter & Haptoglobin & Cefoxitin inter & Valganciclovir inter & Ganciclovir Sodium inter\\Basiliximab inter & Amoxicillin/clavulanic acid inter & HFOV Amplitude & HFOV Frequency & CT Abdomen Pelvis\\Oxacillin inter & Prednisolone inter & Lisinopril inter & Complement C3 Serum & Complement C4 Serum\\Carbamazepine inter & Linezolid inter & Rifampin inter & Pentobarbital inter & Propranolol HCl inter\\Milrinone cont & Azathioprine inter & Octreotide Acetate cont & Nitroprusside cont & Cisatracurium cont\\CSF Bands Percent & Dornase Alfa inter & Bumetanide inter & Terbutaline cont & Allopurinol inter\\Phenytoin inter & Digoxin inter & Cyclophosphamide inter & Calcium Chloride cont & Naloxone HCL inter\\Ketamine cont & Levofloxacin inter & Isoniazid inter & Cisatracurium inter & Norepinephrine cont\\Penicillin G Sodium inter & Factor VII inter & Erythromycin inter & Dobutamine cont & Carvedilol inter\\Labetalol cont & Sodium Bicarbonate cont & Nitrofurantoin inter & Phenylephrine HCl cont & Vitamin E inter\\Haloperidol inter & Esmolol Hydrochloride cont & Aminocaproic Acid inter & Calcium Gluconate cont & Calcium Glubionate inter\\Warfarin Sodium inter & Amphotericin B inter & Metolazone inter & Pentobarbital cont & Doxycycline Hyclate inter\\Atenolol inter & Cyclosporine inter & Amiodarone cont & Doxorubicin inter & Lidocaine cont\\Aminocaproic Acid cont & Isoproterenol cont & Amiodarone inter & Morphine cont & Nitroglycerine cont\\Vecuronium cont & Etomidate inter & Captopril inter & Naproxen inter & Alprostadil cont\\Bumetanide cont & Nesiritide cont & Protamine inter & Sildenafil cont & Procainamide cont\\Flecainide Acetate inter & Acetaminophen/Oxycodone inter & Itraconazole inter & Tacrolimus cont & Infliximab inter\\Cefuroxime inter & Alteplase cont & Cromolyn Sodium inter &  & 
    \end{tabular}
}
\end{table*}
\pagebreak

\section{Learned Binary Mask Algorithmic Description}
\label{section:appendix:b}
\begin{algorithm}[htbp!]
\SetAlgoLined

\textbf{Step 1:} Optimize for intermediate mask $\boldsymbol{m}_{1:T}$ leveraging back-propagation \\

\smallskip
\hspace{5mm} \textbf{Inputs:} $\boldsymbol{x}_{1:T} \in \mathbb{R}^{N \times T}$: dt-patient-matrix defined in Section \ref{section:data:dt} \\
\hspace{5mm} \hspace{11mm} $f(\Theta)$: RNN with trained weights $\Theta$ described in Section \ref{section:methods:rnn} \\

\smallskip
\hspace{5mm} \textbf{Output:} $m_{1:T} \in \mathbb{R}^{N \times T}$: Intermediate mask, solution to Equation \ref{equation:m_float_loss_opt_2}\\

\smallskip
\begin{enumerate}
    \item Generate patient's ROM prediction:  $\boldsymbol{y}_{1:T}=f(\Theta, \boldsymbol{x}_{1:T})$
    \item Modify the RNN by inserting a custom layer, MaskedPerturbLayer, immediately after the input layer
    \begin{itemize}
        \vspace{-0.15cm}\item with weight $\boldsymbol{m}_{1:T} = \sigma(A \times \boldsymbol{z}_{1:T}) \in \mathbb{R}^{N \times T}$, with $\boldsymbol{z}_{1:T}$ initialized to all ones, $\sigma(x) = \frac{1}{1 + e^{-x}}$, and $A=5$
        \vspace{-0.15cm}\item input is $\boldsymbol{x}_{1:T}$ 
        \vspace{-0.15cm}\item output is $\boldsymbol{m}_{1:T} \odot \boldsymbol{x}_{1:T}$
    \end{itemize}
    \item Define loss function for optimization of $\boldsymbol{m}_{1:T}$ as defined in Equation \ref{equation:m_float_loss_opt_2}
    \begin{itemize}
        \vspace{-0.15cm}\item with $\lambda_1=0.005$ and $\lambda_2 = 0.5$
        \vspace{-0.15cm}\item "Freeze" training of weights of all other layers, $\Theta$, such that only weight updated is $\boldsymbol{z}_{1:T}$
    \end{itemize}
    \item Minimize loss function using optimizer RMSProp with initial learning rate of $0.1$
    \begin{itemize}
        \vspace{-0.15cm}\item reduce current learning rate by a factor of $10$ if loss has not been improved over the last $5$ iterations
        \vspace{-0.15cm}\item terminate training if max iteration = $5000$, or if learning rate has been reduced $3$ times
    \end{itemize}
    \item Get intermediate mask from weight of MaskedPerturbLayer: $\boldsymbol{m}_{1:T} = \sigma(A \times \boldsymbol{z}_{1:T})$
\end{enumerate}

\textbf{Step 2} Brute-force grid search to find threshold mask $\boldsymbol{\eta}_{1:T}$ \\

\smallskip
\hspace{5mm} \textbf{Inputs:} $\boldsymbol{x}_{1:T} \in \mathbb{R}^{N \times T}$: dt-patient-matrix defined in Section \ref{section:data:dt} \\
\hspace{5mm} \hspace{11mm} $f(\Theta)$: RNN with trained weights $\Theta$ described in Section \ref{section:methods:rnn}\\
\hspace{5mm} \hspace{11mm} $\boldsymbol{m}_{1:T}\in \mathbb{R}^{N \times T}$: Intermediate mask, solution to \textbf{Step 1} \\

\smallskip
\hspace{5mm} \textbf{Outputs:} $\boldsymbol{\eta}_{1:T} \in \mathbb{R}^{N \times T}$: Threshold mask \\
\hspace{5mm} \hspace{13mm} $\boldsymbol{M}_{1:T} \in \mathbb{R}^{N \times T}$: Binary mask, solution to Equation \ref{equation:m_final} \\

\smallskip
\hspace{5mm} \textbf{Initialize:} $\boldsymbol{\eta}_{1:T} = \{0.5\}^{N \times T}$; $L_{\text{min}} = \infty$; $\lambda_1^2 = 0.0001$; $s_{\text{min}} = 0.05$; $\text{max\_iter} = 2$\\

\smallskip
\For{$i \gets 0$ \KwTo \text{max\_iter}}{

    \For{$t=T$ \KwTo $0$ \text{of each time-step}}{
        Let $u$ be an array containing all unique values in $m_t$ \tcp*{intermediate mask at $t^{th}$ timestep}
        \For{$u\in U$}{
            $\boldsymbol{q}_{1:T} = \boldsymbol{\eta}_{1:T}$ \tcp*{q is used as a temporary threshold} 
            $\boldsymbol{q}_t = u$ \tcp*{$t^{th}$ timestep to a single threshold value}
            $l = f\left(\Theta; \boldsymbol{x}_{1:T} \odot \left(\boldsymbol{m}_{1:T} > \boldsymbol{q}_{1:T} \right) \right) + \lambda_{1}^{2} \norm{1 - \left(\boldsymbol{m}_{1:T} > \boldsymbol{q}_{1:T} \right)}$
            
            \If{$l < L$}{
                $L = l$ \\
                $\boldsymbol{\eta}_{1:T} = \boldsymbol{q}_{1:T}$
            }
        }
    }
    $\boldsymbol{s}_{1:T} = f\left( \Theta; \boldsymbol{x}_{1:T} \odot \left( \boldsymbol{m}_{1:T} > \boldsymbol{\eta}_{1:T} \right) \right)$ \\
    \If{$\forall t \in [0, T]$, $s_t < s_{min}$}{
        terminate
    }
}

\caption{Generating a Learned Binary Mask for an Individual Patient}
\label{alg:lbm}
\end{algorithm}
\pagebreak

\section{Interpreting Individual Predictions: Additional Cases}
\label{section:appendix:c}
\begin{figure*}[htbp!]
\centering
\includegraphics[width=1.0\textwidth]{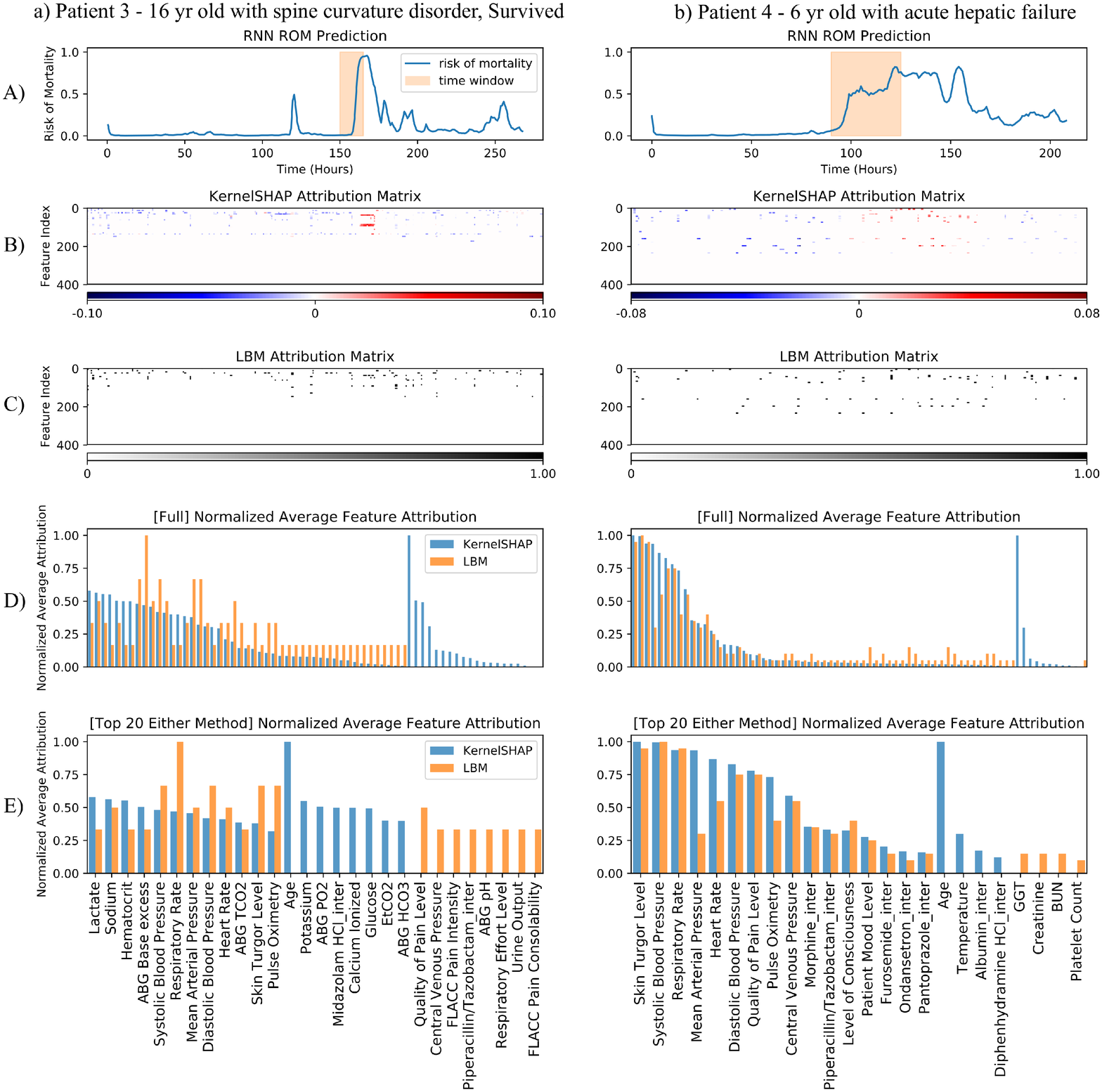}
\caption{Predictions and explanations for two additional individual encounters: a) $p_3$ and b) $p_4$. ROM predictions are visualized in A (top panel). KernelSHAP and LBM attribution matrices are shown as heatmaps in panels B and C, respectively. For KernelSHAP, the heatmap values range from negative to positive in probability units. For LBM, the heatmap is binary.
Attribution matrices were averaged over time periods highlighted in panel A using Equations {\ref{equation: avg_timewindow}} \& {\ref{equation: avg_timewindow_b}} to identify the features that contributed to the increasing ROM predictions in the highlighted time window. These features are visualized in panel D. Finally, panel E visualizes a subset of the features in panel D, presenting only the top 20 features from either method. Note that attribution matrices in B \& C are plotted with time on the x-axis (corresponding to ROM plots in A) and features on the y-axis. Also note that there could be more than 20 variables in E as the selected top 20 features overlap between the methods.}

\label{fig:indiv_plots_appendix}   
\end{figure*}

KernelSHAP and LBM were used to interpret the RNN ROM predictions for two additional patients, and results are presented in Figure {\ref{fig:indiv_plots_appendix}}. In contrast to the two non-surviving patients analyzed in Section {\ref{section:results:indiv}}, the two patients here survived their ICU stay and do not have any illnesses related to respiratory or cardiovascular systems. The first patient, $p_3$, was a 16 year old male with diagnoses of spine curvature disorder, cerebral palsy, and mental retardation. The second patient, $p_4$, was a 6 year old female with diagnoses of acute hepatic failure and chronic pancreatis. Similar to results presented in Section {\ref{section:results:indiv}}, the two patients' RNN ROM predictions were analyzed by averaging their LBM and KernelSHAP attribution matrices corresponding to periods of increasing ROM. These periods are highlighted by the beige window shown in Figure {\ref{fig:indiv_plots_appendix}}A. As in Section {\ref{section:results:indiv}}, Equation {\ref{equation: avg_timewindow}} was applied to obtain $\overline{\boldsymbol{a}}^{p_3}  = \sum_{t=150}^{165} \boldsymbol{a}^{p_3} (t) / (165 - 150)$ and $\overline{\boldsymbol{a}}^{p_4} = \sum_{t=90}^{125} \boldsymbol{a}^{p_4} (t) / (125 - 90)$ for the two patients, and these were subsequently normalized using Equation {\ref{equation: avg_timewindow_b}}.

The top 20 features from both methods during the highlighted time windows of interest are presented in Figure {\ref{fig:indiv_plots_appendix}}E. For $p_3$, the top features from both methods were vitals and lab values. It is worth noting that no importance was placed on respiratory-related measurements (e.g. EtCO2) or infection-related measurements (e.g. Temperature) which constitute a majority of the illnesses and treatments in the pediatric ICU. Interestingly, KernelSHAP and LBM highlighted that Lactate contributed to the increase in ROM predictions of patient $p_3$. Lactate measurements before and during the time window of interest increased (average Lactate before and during: $9.2$ to $24.5$ {mg/dL}). 
What is interesting about these two patients is that these two patients survive and that there is less specificity to their diagnoses than the two patients presented in Section {\ref{section:results:indiv}} who did not survive their ICU stay. While the features highlighted here are not specific to their diagnoses, many of the important features that are highlighted are those of general critical illness which are common to many severity of illness scores \mbox{\citep{pollack1996prism, slater2003pim2}}. Although abnormal levels of Lactate do not directly relate to $p_3$'s diagnoses, it is reassuring that Lactate was highlighted as being important in both LBM and KernelSHAP, as such levels clearly indicate patient decomposition. 

\end{document}